\documentclass[10pt,twocolumn,letterpaper]{article}

\usepackage[pagenumbers]{cvpr} 

\usepackage[dvipsnames]{xcolor}

\definecolor{cvprblue}{rgb}{0.21,0.49,0.74}
\usepackage[pagebackref,breaklinks,colorlinks,citecolor=cvprblue]{hyperref}

\usepackage{xcolor}         
\usepackage{multirow}
\usepackage{makecell}
\usepackage{bm}
\usepackage{pifont}
\newcommand{\xmark}{\ding{55}}%

\def\paperID{15116} 
\def\confName{CVPR}
\def\confYear{2024}

\title{EventSleep: Sleep Activity Recognition with Event Cameras}

\author{Carlos Plou$^{1}$
~~~~~
Nerea Gallego$^{1}$
~~~~~
Alberto Sabater$^{1}$
~~~~~
Eduardo Montijano$^{1}$
~~~~~
Pablo Urcola$^{2}$\\
~~~~~
Luis Montesano$^{1, 2}$
~~~~~
Ruben Martinez-Cantin$^{1}$
~~~~~
Ana C. Murillo$^{1}$\\
{\small $^{1}$DIIS-I3A, Universidad de Zaragoza, Spain ~~~~~ \small $^{2}$Bitbrain Technologies, Spain}
}

\begin{document}
\maketitle

\begin{abstract}
Event cameras are a promising technology for activity recognition in dark environments due to their unique properties. 
However, real event camera datasets under low-lighting conditions are still scarce, which also limits the number of approaches to solve these kind of problems, hindering the potential of this technology in many applications. 
We present \texttt{EventSleep}, a new dataset and methodology to address this gap and study the suitability of event cameras for a very relevant medical application: sleep monitoring for sleep disorders analysis. 
The dataset contains synchronized event and infrared recordings emulating common movements that happen during the sleep, resulting in a new challenging and unique dataset for activity recognition in dark environments. Our novel pipeline is able to achieve high accuracy under these challenging conditions and incorporates a Bayesian approach (\textit{Laplace ensembles}) to increase the robustness in the predictions, which is fundamental for medical applications. Our work is the first application of Bayesian neural networks for event cameras, the first use of  \emph{Laplace ensembles} in a realistic problem, and also demonstrates for the first time the potential of event cameras in a new application domain: to enhance current sleep evaluation procedures. 
Our activity recognition results highlight the potential of event cameras under dark conditions, and its capacity and robustness for sleep activity recognition, and open problems as the adaptation of event data pre-processing techniques to dark environments.
\footnote{Links to data and code at: \url{https://sites.google.com/unizar.es/eventsleep/home}}
\end{abstract}

\section{Introduction}
\label{sec:intro}
\begin{figure*}[!htb]
     \includegraphics[width=\linewidth]{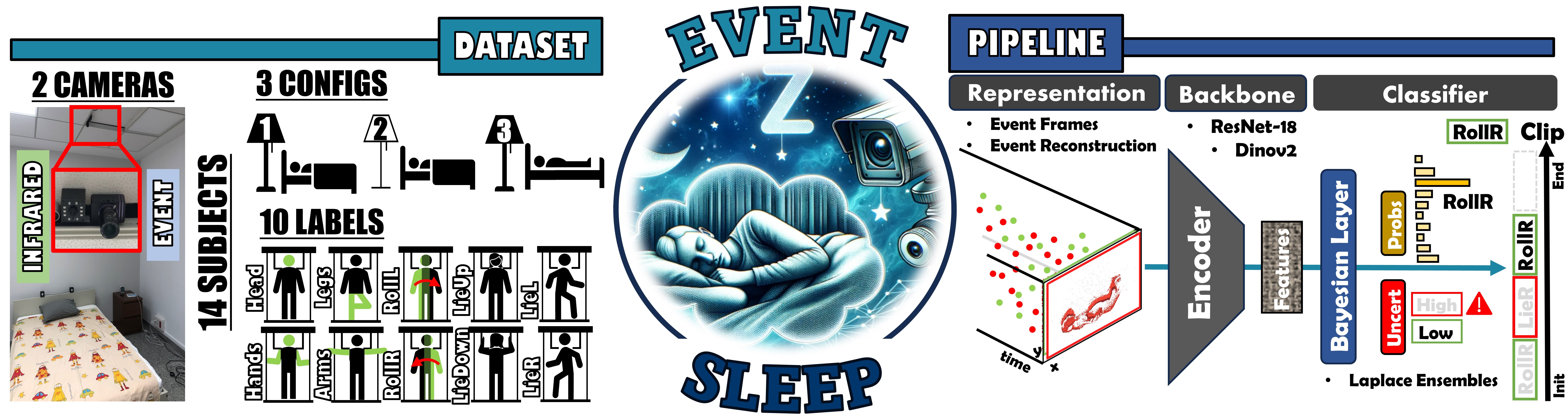}
     \caption{Overview of \texttt{EventSleep}. The left part summarizes the dataset, showing the room used to record, the three configurations considered, depending on the blanket and bedlamp state, and the ten sleep activity classes labeled. The right part shows the stages of the pipeline presented. Our approach for activity recognition from event data includes a Bayesian approach that achieves the best performance and robustness to noise and misclassification errors.}
     \label{fig:setup}
 \end{figure*}
Sleep occupies a significant portion of our lives, roughly one third of our lifetime. Unfortunately, sleep disorders have become increasingly prevalent, with up to 50$\%$ of the adult population reporting problems related to their sleep~\cite{vanderlinden2020effects}. Many sleep disorders, including restless legs syndrome, periodic limb movement in sleep, somnambulism, or bruxism, are characterized by abnormal movements during sleep. However, the diagnosis of most of the sleep-related pathology is bottle-necked by sleep scoring based on polysomnographpy, which has traditionally relied on manual classification by expert sleep technicians~\cite{hassan2017decision, huang2014knowledge}. 

This work is part of a large project centered on the study of sleep and our concern was whether event-based sensors could yield valuable insights into this field. Specifically, event cameras could provide automatic information about a subject's actions during sleep -i. e. \emph{sleep activity recognition}- in order to complement the information provided by other sensors (EEG, ECG, SpO2, etc.) which can actually be perturbed by body motions. 
The primary obstacle cameras encounter in sleep activity recognition is darkness. In response to this challenge, researchers have already explored the utilization of near-infrared and depth cameras. Differently, we explore the suitability of event cameras which (1) can also work under low lighting, (2) provide higher temporal resolution, (3) offer an increase of privacy of the recorded subject with respect to infrared recordings, as detailed in ~\cref{sec:events_representation}, and (4) enable low power consumption (it consumes power only when events occur, and subjects are expected to exhibit sparse and intermittent actions throughout the night). However, these properties had never been tested under low-light conditions, since there is no event-based public dataset recorded under darkness for complex tasks as activity recognition.

In this work, we present a dataset, \texttt{EventSleep}, that demonstrates the opportunities that event cameras bring in dark lighting conditions, and in particular, to explore activity recognition during the sleep. 
Experts participating in sleep study projects have identified specific actions, movements or subject positions, that would be interesting to recognize to help with their studies, and correspond with the ten classes provided in the dataset. 
\texttt{EventSleep} comprises event-data recordings and annotations across three distinct scenarios, with varying lighting and bed sheet conditions, and synchronized infrared image recordings of the same sequences. \Cref{fig:setup} shows an overview of the dataset and the setup. We also present a novel event-based action recognition pipeline that tackles the dataset challenges. It outperforms relevant baselines, including the state-of-the-art for activity recognition with event data. 

Medical applications such as sleep disorder diagnosis require robust tools that are aware of perturbations and uncertainty sources. A common problem of many deep learning approaches is the lack of uncertainty quantification and suffer from hallucination in out-of-distribution data, which hinders the reliability of the predictions. Bayesian neural networks can be used to quantify both model (epistemic) uncertainty and data (aleatoric) uncertainty. Our pipeline also incorporates an uncertainty quantification method based on Bayesian deep learning, improving robustness and reliability in posterior decision making in the study of sleep disorders.
Overall, the contributions of this work are: 
\begin{itemize}
    \item A unique and challenging event-based dataset  focused on activity recognition in dark environments.
    \item A novel approach for activity recognition in dark environments, featuring calibrated uncertainty estimation via \emph{Laplace ensembles}. This is the first evaluation of  \emph{Laplace ensembles} in a real-world computer vision scenario.
    \item The first proof of concept for a new event camera application: recognition of relevant actions for sleep pathologies study. 
\end{itemize}

\section{Related Work}

\textbf{Sleep activity recognition.}
Sleep activity recognition has been studied with different sensory inputs.  
Most of the related literature relies on wearable sensors \cite{yadav2021review,zhang2019human} or environmental sensors such as sound or light sensors \cite{kay2012lullaby}. 
The majority of datasets that use wearable sensors are based on polysomnograms, which is the standard for sleep quality measurements. However, the process of gathering the physiological signals of a polysomnogram, such as an electroencephalogram (EEG), requires the subjects to be monitored by a fully equipped unit with technicians. There are several public datasets using this kind of setup \cite{kemp2000analysis,terzano2001atlas,khalighi2016isruc}, but there is limited prior work with external sensors, despite their interest for unimodal or multimodal activity recognition \cite{yadav2021review,sathyanarayana2018vision}, when complemented with other wearable sensors. 
When it comes to non-invasive sensors, video data is the principal data modality used. One of the main challenges of preliminary studies on image analysis for sleep monitoring is the low light condition \cite{nakajima2000monitor}. To overcome this challenge, researchers have already investigated the use of near-infrared and depth cameras. Initially, Mohammadi \etal \cite{mohammadi2020transfer} employed a single infrared (IR) camera to capture 11 sleep behaviors in 12 participants concurrently with PSG. Additionally, the BlanketSet dataset comprises RGB-IR-D recordings of 8 sleep actions by 14 subjects in a hospital bed \cite{carmona2023blanketset}. Both datasets involve non-clinical healthy participants to enable data sharing and robust labeling, as in our case.
The closest alternative to cameras is the use of pressure sensors on the bed, which provides a \emph{heat map} of the body of the subject \cite{matar2019artificial}. Differently from all these works, our dataset primary sensor is an event camera.

\noindent
\textbf{Event-based datasets for activity recognition.}
Since event cameras present high robustness to challenging lighting conditions and fast motion, we find different event-based datasets designed for different action recognition applications. \Cref{tab:datasets} shows a summary of the most relevant public benchmarks. 
Most of them are designed for recognizing general daily actions, with the exception of ASL-DVS \cite{bi2020graph} and SL-Animals-DVS \cite{vasudevan2021sl}, which are specialized for sign language gesture recognition. 
The number of classes spans from 5 to 50 (median 12), with the number of subjects ranging from 5 to 105 (median 18). The datasets resolution extends from 128 $\times$ 128 pixels up to 346 $\times$ 260 pixels, with the notable exception of  THU$^{\text{E-ACT}}\text{-50}$ \cite{gao2023action} which boasts a resolution of 1280 $\times$ 800 pixels. It stands out as the most comprehensive dataset for general action recognition, surpassing others in terms of resolution, size, subjects, and the number of classes.

\begin{table*}[!tb]
    \centering
    \caption{\texttt{EventSleep} compared to related event-based activity recognition datasets. 
    }
    \resizebox{\textwidth}{!}{%
    \begin{tabular}{l l c c c c c c c c c}
    \toprule
        Dataset & Activity & Modality & Resolution &  Classes &  Subj. & Dark  & Occ. & Clips & Time/Clip & Total Time\\
    \midrule
        n-HAR \cite{pradhan2019n} & General & Events & $304 \times 240$ & 5 & 30 & No & No & 3,091 & N/A & N/A  \\
        
        DailyAction \cite{liu2021event} & General & Events & $128 \times 128$ & 12 & 15 & No & No & 1,440 & N/A & N/A \\
        
        DVS128 Gesture \cite{amir2017low} & General 
        & Events+RGB & $128 \times 128$ & 11  & 29 & No & No & 1,342 & 6s & 8,052s \\
        
        THU$^{\text{E-ACT}}\text{-50}$ \cite{gao2023action} & General & Events+RGB & $1280 \times 800$ & 50 & 105 & No & No & 10,500 & 3.5s & 36,750s \\
        
        THU$^{\text{E-ACT}}$\text{-50-CHL} \cite{gao2023action} & General & Events+RGB & $346 \times 260$ & 50 & 18 & * & No & 2,330 & 3.5s & 8,155s \\
        
        PAF \cite{miao2019neuromorphic} & Office & Events & $346 \times 260$ & 10 & 15 & No & No & 450 & 5s & 2,250s \\
        
        ASL-DVS \cite{bi2020graph} & SL
        & Events & $240 \times 180$ & 24  & 5 & No  & No & 100,800	& 0.1s & 10,800s \\
        
        SL-Animals-DVS \cite{vasudevan2021sl} & SL 
        & Events & $128 \times 128$ & 19  &  59 & No & No & 1,121 & 4.26s & 4,775s \\
        \midrule

        \texttt{EventSleep} & \textbf{Sleep} 
        & \textbf{Events+IR} & $640 \times 480$  & 10 & 14 & \textbf{Yes} & \textbf{Yes} & 1,016 & 5.07s & 5,151s \\
    \bottomrule    
    \multicolumn{11}{l}{*THU$^{\text{E-ACT}}\text{-50-CHL}$ includes some poorly illuminated clips, but not in low light or dark scenarios as \texttt{EventSleep}.}
    \end{tabular}}
    \label{tab:datasets}
\end{table*}

Although some of these datasets are recorded with different light sources, only THU$^{\text{E-ACT}}\text{-50-CHL}$ \cite{gao2023action} presents challenging illuminations, but no dark or near-dark scenes. Existing event camera datasets with low lighting recordings are mostly devoted to driving scenarios~\cite{zhu2018multivehicle, perot2020learning} and still present external illuminations such as street lamps or car lights. Therefore, there is a scarcity of datasets that include sequences without proper illumination. This hinders the potential of event cameras for people monitoring in dark scenarios and their suitability for applications such as sleep monitoring. 
For this purpose, we present \texttt{EventSleep}, a challenging dataset for sleep activity recognition which, as highlighted in \cref{tab:datasets}, is the first one recorded in dark conditions and including significant human body occlusions.

\noindent
\textbf{Event-based methods for activity recognition.}\label{sec:related_ev_methods}
Deep learning is the common choice for event-based activity recognition. 
Among the approaches for recognition from event data, we find models designed to benefit from the event data sparsity by using PointNet-like Networks \cite{wang2019space}, Graph Neural Networks \cite{bi2020graph, deng2021ev}, or Spike Neural Networks 
\cite{shrestha2018slayer, parameshwara2021spikems}. Most approaches transform sets of incoming events into dense event representations, or frame representations, before further processing. 
Examples of these representations include
time-surfaces~\cite{lagorce2016hots}, surfaces of Active Events~\cite{mueggler2015lifetime}, binary frame representations~\cite{ghosh2019spatiotemporal, innocenti2021temporal}, histograms~\cite{Sabater_2022_CVPR}, queing mechanisms~\cite{baldwin2022time} or learned models~\cite{cannici2020differentiable}.
Dense frame representations are commonly processed with Deep Learning models such as CNNs \cite{amir2017low, cannici2020differentiable, innocenti2021temporal, baldwin2022time} or Transformers \cite{sabater2023event, peng2023get}.
Additionally, some methods do not analyze individual frame representations but sets of them, typically processing each frame independently and then aggregating the intermediate results with RNNs \cite{innocenti2021temporal, weng2021event}, CNNs \cite{amir2017low, innocenti2021temporal}, or attention-based methods \cite{sabater2023event, zhang2022spiking, wang2022exploiting}. 
 Recently, a novel event representation, Group Token, is introduced to train an event-based ViT backbone called Group Event Transformer (GET). In our pipeline, we build dense event representations with FIFOs, similar to Sabater \etal \cite{sabater2023event}.

\noindent
\textbf{Uncertainty quantification with Bayesian deep learning.}
These approaches perform probabilistic inference on deep network models, enabling uncertainty quantification. In high dimensional deep networks, Bayesian inference is intractable and rely on some approximate inference methods, e.g., variational inference \cite{Osawa2019} or Laplace approximations \cite{mackay1992bayesian}, approximate the posterior distribution by a mean-field Gaussian distribution, which might perform poorly in high dimensions. In practice, sampling methods based on bootstrapping, like deep ensembles \cite{Lakshminarayanan2017}, or Monte Carlo, like MC-dropout \cite{Gal2016}, have shown to perform better for very deep networks \cite{Gustafsson2019}, but they require large number of samples and still suffer from uncertainty calibration. The closest to our approach is Eschenhagen \etal \cite{eschenhagen2021mixtures}, which presented a method combining both techniques (bootstraping and Laplace), but that work only demonstrates clear benefits in toy problems (MNIST, CIFAR-10) with artificially corrupted data. 

\section{\texttt{EventSleep} Dataset}\label{sec:dataset}

The \texttt{EventSleep} dataset is a valuable benchmark for event cameras under low light conditions. 
To ensure the dataset's utility and reproducibility, several crucial factors were considered, including the variety of activities and its relation with some sleep disorders, the recording methodology, and the scene configurations involved. These considerations are discussed next. 
Although the primary aim of \texttt{EventSleep} is to provide researchers with a novel and unique resource for sleep activity recognition and analysis, we have also designed \texttt{EventSleep} as a challenging benchmark that allows, in general, the evaluation of event data processing in dark environments with strong occlusions of the subjects recorded.

\subsection{Design and Methodology}
\label{sec:design}
\noindent
\textbf{Cameras.}
Trials were recorded by two different cameras: Event (DVXplorer camera, 640 x 480 resolution) and Infrared (ELP HD Digital Camera). 
Both cameras were placed together, attached to a metal bar, facing down and located 2 meters above the center of the bed, ensuring a full observation of the participants movements (\cref{fig:setup}). Furthermore, we employ a light meter during the recordings to precisely measure the ambient light levels surrounding the subject's face (it has a  measurement range from $0.1$Lux to $200,000$Lux). Both camera recordings are synchronized.

\noindent
\textbf{Recording procedure.} 
It is important to notice that, just as previous sleep activity recognition datasets \cite{mohammadi2020transfer, carmona2023blanketset}, \texttt{EventSleep} is not a real medical case. In other words, our recordings are not actual sleep activity recordings of real patients. The reason is twofold. First, since subjects are volunteers and are not part of a clinical trial, we have the ability to release all the data, enriching the scientific community. Second, volunteers were instructed to perform a sequence of activities, to facilitate obtaining ground truth labels and a reasonably balanced dataset for all classes.
The participants were instructed about the main actions present in sleep disorders and the goal of our study. 
We specified to all subjects an ordered list of actions they should perform, but we deliberately avoided providing many details or specific examples, and we asked them to perform the movements as naturally as possible. This approach was taken aiming to capture a broad diversity of behaviors. We address ethical considerations regarding participants in Sec.~\ref{Sec:SuppDataset} of the supplementary material.

\noindent
\textbf{Scene configurations.}
\texttt{EventSleep} was recorded in a room that mimics a regular bedroom. Each participant was recorded under 3 settings:
\begin{itemize}
\item  \textit{Config1}: The subject is covered by an eiderdown under full darkness.
\item  \textit{Config2}: The subject is covered by an eiderdown under partial darkness.
\item  \textit{Config3}: The subject is uncovered under full darkness.
\end{itemize}

\noindent The full darkness setup ($\leq 0.1$Lux) consists of a bedroom with all lights off, door and window closed, with dense blinds covering the window. 
The partial darkness setup ($0.2$Lux) has a small night lamp on the floor, away from the subject head.

\noindent
\textbf{Activities.}
All subjects were instructed to execute a certain set of movements and positions during each trial. These movements were carefully selected based on their potential relevance to medical research as part of a larger biomedical study project. 
Within this context, experts in the field identified specific actions that appear in normal sleep (e.g. transitions between sleep positions) and in sleep disorders such as restless legs syndrome or periodic limb movement \cite{pavlova2019sleep} (e.g. shaking the legs). Since many sleep studies require the recording of physiological information, we also considered positions that could impact the performance of other sensors such as a headband for polysomnography.

\begin{figure*}[!tb]
\centering
\includegraphics[width=0.9\linewidth]{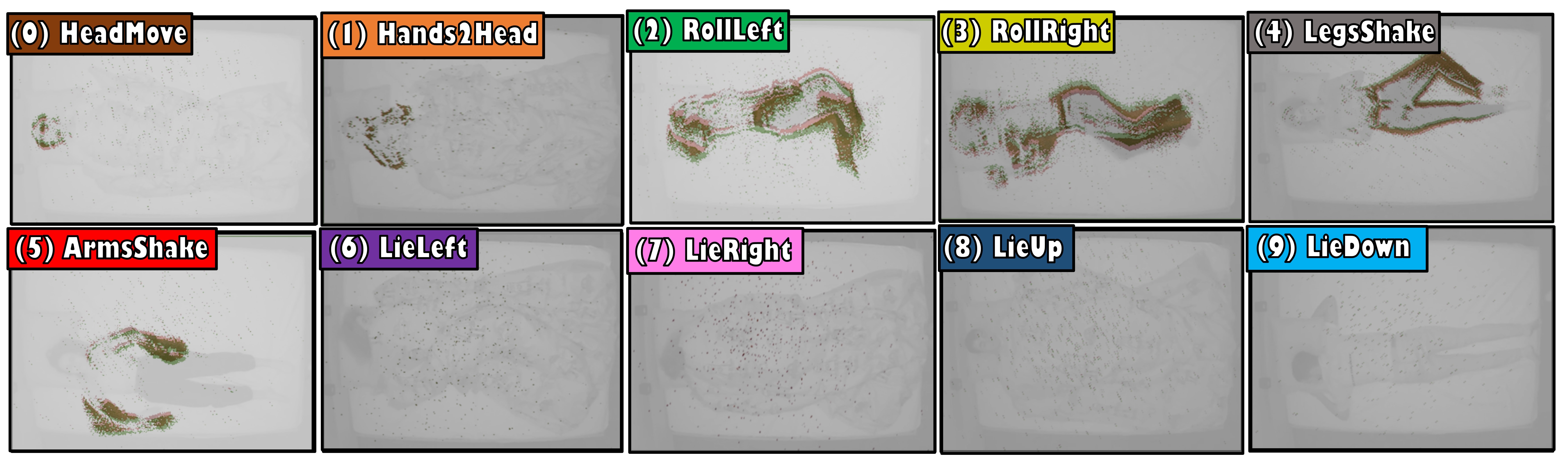}
    \caption{\texttt{EventSleep} classes. Each plot shows an event frame (red and green pixels) of each class superposed to the corresponding infrared frame (grayscale).}
    \label{fig:Labels}
\end{figure*}

The final set of \textbf{10 activity labels} considered is: 
(0) \textit{HeadMove}: adjustments or repositioning of the head. (1) \textit{Hands2Face}: touching or placing the hands on the face or head.
(2) \textit{RollLeft}: rolling the body towards his/her left. (3) \textit{RollRight}: rolling the body towards his/her right. (4) \textit{LegsShake}: shaking or moving the legs.
(5) \textit{ArmsShake}: shaking or moving the arms.
(6) \textit{LieLeft}: lying in a left lateral recumbent position. (7) \textit{LieRight}: lying in a right lateral recumbent position.
(8) \textit{LieUp}: lying in a supine position.
(9) \textit{LieDown}: lying in a prone position. 
 \cref{fig:Labels} shows an example of an event frame (following the representation detailed in next section~\cref{sec:approaches}) for each class.

\noindent
\textbf{Challenges.} 
The presented dataset poses several interesting challenges for related research. (1) Darkness implies recording with high sensitivity, which produces \textbf{high levels of noise}. This difficulties some event pre-processing techniques, such as denoising. Besides, 
the state-of-the-art for event image reconstruction under low-light scenarios~\cite{Rebecq19pami} struggles to process our data, and it is not possible to directly identify subjects, as illustrated with the example in \cref{fig:IRvsReconst} and the additional examples in the supplementary material Sec.~\ref{Sec:SuppConfig}. 
(2) Four of the activity labels considered correspond to poses during the sleep. During this \textbf{almost static classes}, since the subject is quiet, theoretically there should not be enough events to distinguish among them.  However, it is impossible to be completely quiet, e.g., subjects continue breathing, and then these clips could contain some residual information which, combined with more temporal context, could be leveraged to classify correctly. 
(3) Ideally, the sleep activity analysis should be performed online, to allow for interventions during the sleep to help with the subject pathologies. This brings an extra challenge due to the \textbf{computational requirements} to run the recognition tasks online.

\begin{figure}[!tb]
    \centering
       \begin{subfigure}{0.45\columnwidth}
        \includegraphics[width=\linewidth]{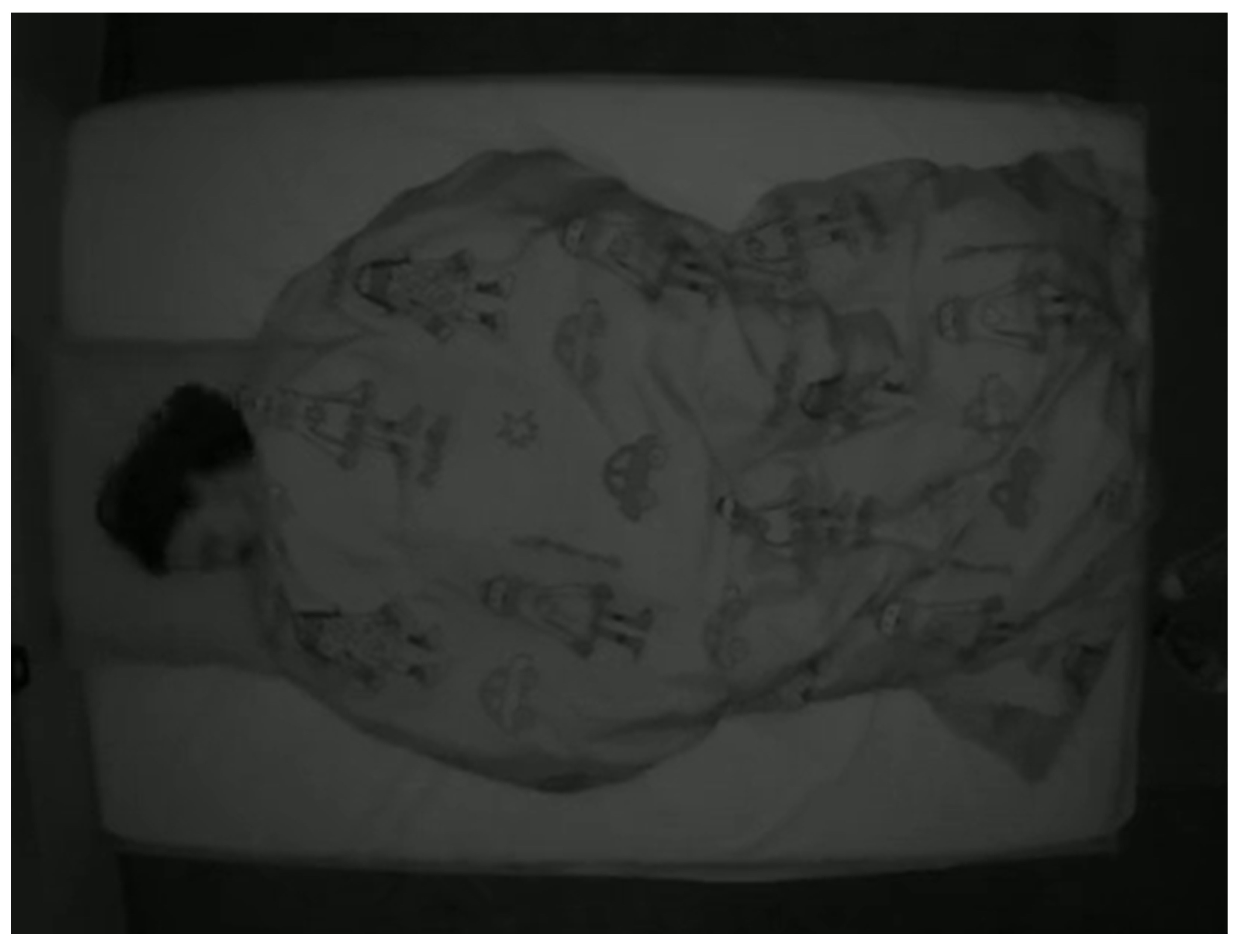}
    \end{subfigure}
    \begin{subfigure}{0.45\columnwidth}
        \includegraphics[width=\linewidth]{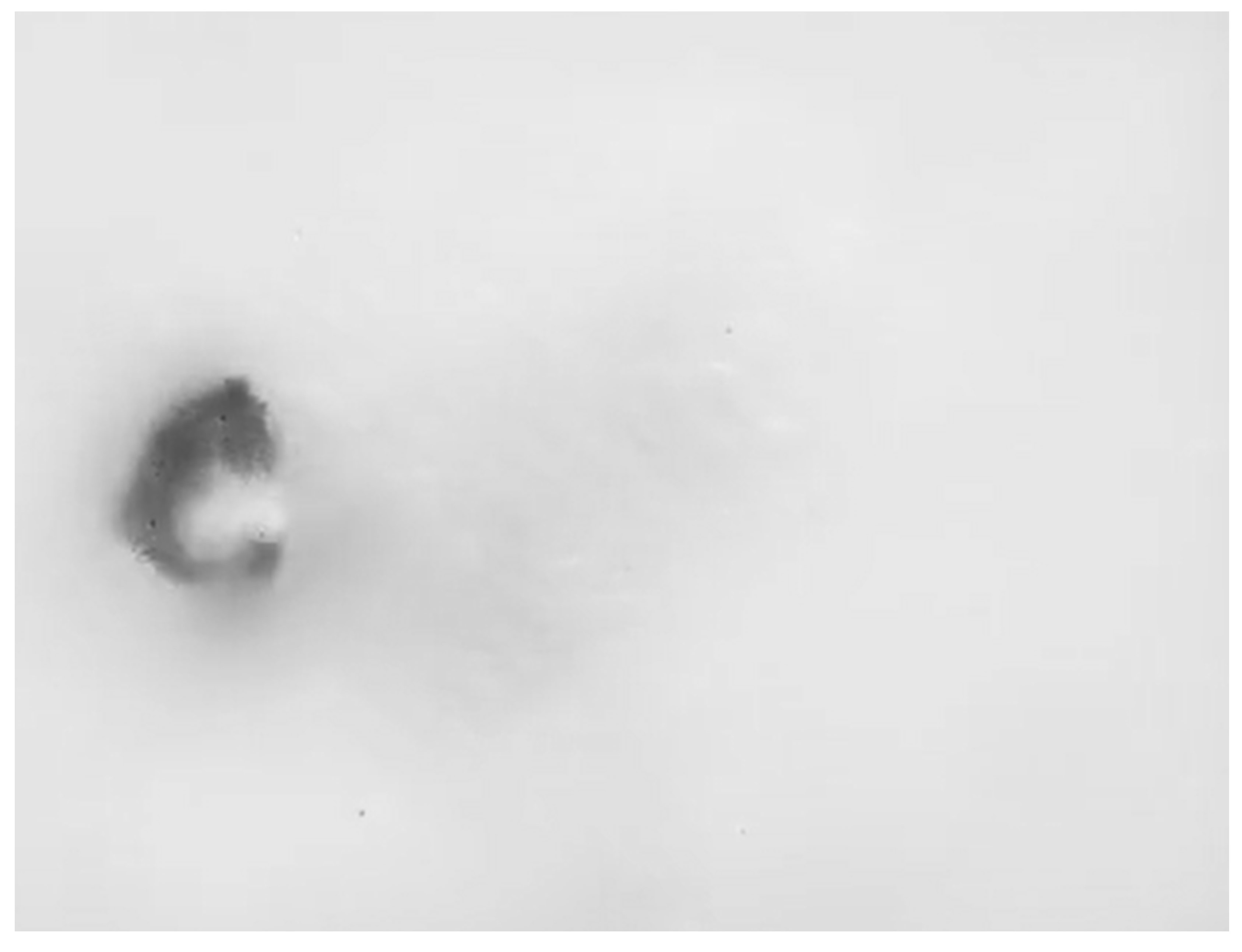}
\end{subfigure}
\caption{Privacy comparison: IR frame (left) vs reconstruction from events~\cite{Rebecq19pami} (right). More examples in Sec.~\ref{Sec:SuppConfig}.}
 \label{fig:IRvsReconst}  
 \end{figure}

\noindent
\textbf{Content and accessibility.}
The dataset consists of a total of 1,016 clips, acquired as part of 42 trials (14 participants under 3 different configurations). Each trial contains over 20 action clips, each lasting about 5 seconds.  There are separate training (744 activity clips, from 10 subjects, 3 trials each) and test (272 activity clips, from 4 other subjects not included in training, 3 trials each) sets. The training and test set were built to increase diversity and minimize biases, such as sex or appearance. 
The test set also includes the untrimmed recordings for each subject, to facilitate online action recognition proposals evaluation.

The Python code repository contains scripts to facilitate the dataset exploitation -e.g., event frames construction, synchronized visualization with IR, labels details- together with a toy dataset that provides a more complete overview of the \texttt{EventSleep} dataset content. The metadata, licenses and more details can be found in the README files of the repository. Links to both the dataset and code repositories are available on \url{https://sites.google.com/unizar.es/eventsleep/home}.

\section{Approach}
\label{sec:approaches}

In contrast to conventional RGB cameras, event cameras record visual data sparsely and asynchronously. When an alteration in intensity occurs, the camera generates an event ($e$), which includes its location ($x,y$) within the sensor grid's space $\left( H \times W\right)$, its timestamp ($t$) measured in microseconds, and its polarity ($p$) to show whether the change is positive or negative. Each clip contains a set of events $\mathcal{E}=\left\{e^{i}|e^{i}=\{x^{i},y^{i},t^{i},p^{i}\}\right\}_{i=1}^{n}$ sorted by its timestamp.

Our proposed event-based action recognition pipeline: 1) addresses the darkness challenge of the dataset leveraging  event pre-processing and representation techniques, 2) exploits existing models for visual tasks using RGB data and, 3) incorporates uncertainty estimation to achieve the robustness demanded by the medical applications. \Cref{fig:setup} depicts the main steps of our pipeline, which are detailed along this section: \textit{Events representation}, \textit{Backbone} used to encode useful features and the final \textit{Classifier}.

\subsection{Events representation.}\label{sec:events_representation}

We represent event data with up to 3-channel frames, following two prior work representations for recognition with events, as detailed next. 

\noindent \textbf{Online Event Frames.}
A common technique to process event data is to transform sparse events into dense frame representations.
The event frame representation we use is based on EvT+ \cite{sabater2023event}. This event representation is obtained by aggregating the event data in frames with a frequency of $\Delta t$. In other words, we build a frame  in the timestamps 
$\left\{t_k | t_k=t^{0}+k\Delta t\right\}_{k=1}^{N}$, where $N=\lfloor \frac{t^{n}-t^{0}}{\Delta t}\rfloor.$ 

Considering timestamp $t_{k}$, for each pixel 
within the sensor array and each polarity, we save the last event detected in the last $T_M$ seconds, i.e., in the temporal window $[t_k-T_M, t_k]$. Afterwards, the event timestamps are used to build the intermediate frame representation $\mathbf{F}_{k} \in \mathbb{R}^{H \times W \times 2}$ such that,%
\begin{equation}
\mathbf{F}_{k}(x,y,p)= 
  t^{i}, \quad \text{if} \hspace{1mm} \exists  e^{i} \in \mathcal{E} \hspace{1mm} s.t. \hspace{1mm}  \begin{cases}
           (x^{i},y^{i},p^{i})=(x,y,p) \\
           t^{i} \in [t_k - T_M, t_k]
         \end{cases}  
\end{equation}

We use \texttt{NaN} for undefined pixel values and normalize and scale the rest\cite{sabater2023event}: 
\setlength{\abovedisplayskip}{3pt}
\setlength{\belowdisplayskip}{3pt}
\begin{equation}
\mathbf{F}_{k}(x,y,p) := \frac{\mathbf{F}_{k}(x,y,p) - (t_k - T_M)}{T_M}.
\end{equation}
Events triggered at the end of the time window will approach $1$, while those at the start will approach $0$. The final set of frames is $\mathcal{F}=\left\{\mathbf{F}_{k}\right\}_{k=1}^{N}$, $\mathbf{F}_{k} \in [0,1]^{H \times W \times 2}$.

\noindent \textbf{Offline Event Reconstruction.} 
In an attempt to boost the classification of the most challenging labels (when there are nearly no events), we suggest to leverage gray-scale reconstruction from events as an additional input channel of the representation. Ercan \etal \cite{ercan2023hypere2vid} recently showed that E2VID \cite{Rebecq19pami} achieves the highest performance in low-light scenarios.  
E2VID runs a convolutional recurrent network (CRN) to reconstruct video from a stream of events. They first represent the events in a spatio-temporal voxel grid $\mathbf{E}\in [0,1]^{H \times W \times B}$. To achieve this, events are discretized into $B$ temporal bins,%
\setlength{\abovedisplayskip}{3pt}
\setlength{\belowdisplayskip}{3pt}
\begin{equation}
    \mathbf{E}(x, y, t)=\sum_{x^{i}=x, y^{i}=y}p^{i} \max(0, 1-|t^{n}-t^{i}_{*}|),
\end{equation}
where $t^{i}_{*}=(B-1)\frac{t^{i}-t^{0}}{t^{n}-t^{0}}$ is the normalized event timestamp. Subsequently, $\mathbf{E}$ is processed in the CRN, yielding a video of $N'$ gray-scale frames $\mathbf{R}_{k} \in [0,1]^{H \times W \times 1}$. The utilization of the event reconstruction channel serves, as well, to verify the level of privacy offered with the event data in our settings. As the examples in ~\cref{fig:IRvsReconst} and in Sec.~\ref{Sec:SuppConfig} clearly suggest, it is much more difficult to recover personal information from event data than from a low cost infrared camera. Also, we found that this technique has difficulties given the extremely noisy nature of our data.

Both representations $\mathbf{F}_{k}, \mathbf{R}_{k}$ can be synchronized and stacked,  getting 3-channel frames $\mathcal{G}=\left\{\mathbf{G}_{k}=\{\mathbf{F}_{k}, \mathbf{R}_{\lfloor r*k \rfloor}\} \right\}_{k=1}^{N}$, where $r=\frac{N'}{N}$.

\subsection{Architecture}\label{Sec:Architecture}
Our architecture is composed by a visual backbone (CNN or ViT) to obtain features and a classifier to process them.

\noindent \textbf{Backbone.}
The described event representation outputs a set of frames for each clip encoding temporal information at $\Delta t$ rate.  In our experiments, we found that the visual information encoded in RGB pretrained encoders can be transferred and exploited by event frames.
We hypothesize that the temporal encoding as pixel value actually helps on transferring features from image classification to action recognition tasks.

\emph{Dinov2-ER}.
We use a frozen visual transformer (ViT) trained using self-supervision in a large set of visual tasks (using RGB data) \cite{oquab2023dinov2}. The original model has 1 billion parameters, but we use a distilled version (ViT-S/14). This approach takes 3-channel frames $\mathcal{G}$, which combine both event frames and reconstruction representations. It returns a set of features $\mathbf{f}_{k} \in \mathbb{R}^{384}$ for each frame.

\emph{ResNet-E}. We fine-tune a ResNet-18~\cite{he2016deep} pretrained on ImageNet, 
using only the 2-channel event frames $\mathcal{F}$. We did not observe improvements adding the reconstruction channel, and using only 2-channels reduces the computational overhead and allows on-line evaluation of frames, instead of per clip. This backbone provides a set of features $\mathbf{f}_{k} \in \mathbb{R}^{1000}$ for each frame.

\noindent \textbf{Classifier.} 
We employ a fully-connected layer followed by a softmax layer to obtain probability scores $\mathbf{p}_{k}\in~[0,1]^{10}$. 
The final clip prediction is computed as the \textit{mode} frame prediction $\hat{y}~=Mode\left(\mathcal{Y}\right)$, where $\mathcal{Y}=~\{ \arg\max_{j} \mathbf{p}_{k}^{j}\}_{k=1}^{N}$.

\subsection{Bayesian deep learning}
In sensible deep learning applications, such as medical and health related ones, it is just as crucial to attain high accuracy as it is to ensure the reliability of the predictions. Deep learning models are known to be overconfident, specially with out-of-distribution data or missmodeling error.  This problem occurs when the model's confidence (defined as $c_k = \max_{j}\mathbf{p}_{k}^{j}$) is overestimated. Probabilistic learning methods, such as Bayesian deep learning replaces the single point of the estimate of the network parameters $\bm{\theta}^*$ by a posterior distribution $p(\bm{\theta}| \mathcal{D})$ conditioned to the training data $\mathcal{D}$. This enables a better estimation of the uncertainty and increased robustness in the predictions, using the posterior predictive distribution $p(\mathbf{y}|\mathbf{x}) = \int p(\mathbf{y}|\mathbf{x},\bm{\theta}) p(\bm{\theta}| \mathcal{D}) d\bm{\theta}$ where $\mathbf{x}$ is the frame input (see \cref{Sec:bayesian}). However, the posterior and predictive posterior distributions are intractable, relying on approximate Bayesian inference.

\noindent \textbf{Laplace approximation.} Mackay \cite{mackay1992bayesian} first introduced the idea of approximating the parameters posterior distribution by a multivariate Gaussian distribution $ p(\bm{\theta}| \mathcal{D})~\approx~\mathcal{N}(\bm{\theta}^*, \mathbf{H}^{-1}):=\mathcal{N}_{\pmb{\theta}^{*}}$, where $(\bm{\theta}^*, \mathbf{H})$ are the optimum and Hessian at that point of the loss function, the negative log posterior (NLP). The intuition of this approach is to approximate the NLP locally at the optimum by a second-order Taylor expansion \cite{mackay1992bayesian}. Unfortunately, computing $\mathbf{H}$ is highly resource-intensive, and it is only tractable for small neural networks or subnetworks \cite{daxberger2021bayesian}. In this case, we compute the posterior distribution to the last layer, using Laplace-Redux \cite{daxberger2021laplace}.
Finally, this posterior distribution over the weights is translated into a predictive classification distribution via Laplace bridge approximation \cite{hobbhahn2022fast}.

\noindent \textbf{Deep ensembles.} 
This method generates a set of samples by training multiple networks from random intializations~\cite{Lakshminarayanan2017}. Strictly speaking, deep ensemble samples are not generated from the full posterior distribution $p(\bm{\theta} | \mathcal{D})$, but the fact that they start from random locations may capture the multimodality of the predictive posterior distribution \cite{Gustafsson2019}. By optimizing each sample on the MAP loss we assure that each sample has high probability, which is fundamental for deep models where the weight space $\bm{\theta}$ is huge, but the number of samples $S$ must remain small for computational tractability. For the experiments, we use $S=32$, while trying to avoid losing the pretraining of the encoder: Dinov2 encoder is frozen and the ensembles apply only to the classifier; for ResNet, the initialization distribution for the pretrained encoder weights has smaller variance than for the classifier weights.

\noindent \textbf{\textit{Laplace ensembles}.} Although Bayesian methods outperform deterministic methods in term of robustness and improve the uncertainty estimation, the curse of dimensionality still induces overconfident results. Inspired by Eschenhagen \etal \cite{eschenhagen2021mixtures}, \textit{Laplace ensembles} reasoning is to combine the global uncertainty captured by the multimodality of deep ensembles with the
local uncertainty around a single mode captured by the Gaussian of Laplace approximation. As result, we obtain a  mixture of Gaussians $\left\{ \mathcal{N}_{\pmb{\theta}^{*}_{1}}, \ldots, \mathcal{N}_{\pmb{\theta}^{*}_{S}} \right\}$ where each of the $S$ ensembles is represented by a Gaussian distribution $\mathcal{N}_{\pmb{\theta}^{*}_{i}}$ using Laplace approximation. Differently from previous work \cite{eschenhagen2021mixtures}, which employs multi-class probit approximation \cite{gibbs1998bayesian} to get the predictive distribution from each sample $\mathcal{N}_{\pmb{\theta}^{*}_{i}}$, we observed higher performance leveraging Laplace bridge approximation \cite{hobbhahn2022fast}.

\noindent \textbf{Probabilistic clip predictions.}
Thanks to the reliable probabilities of our predictions, we can leverage the uncertainty that lie inside these probabilities to get the prediction of the clip. Instead of taking the \textit{mode} among all frames prediction as before, we predict from the \textit{accumulated probability} $\hat{y}=~\operatorname{argmax}_{j} \sum_{k=1}^{N}\mathbf{p}_{k}^{j}$.

\section{Experiments}\label{sec:case}
The following experiments show promising results with the presented approaches for sleep activity recognition, and the open challenges of \texttt{EventSleep} dataset. 
\subsection{Baselines.}
The potential of the proposed \texttt{EventSleep} dataset for sleep monitoring, and understanding its challenges, is evaluated by comparing our results with three different baseline methods of different levels of complexity:  
\textit{Blobs-E}, \textit{GET}, \textit{ResNet-IR}. 
They are chosen, respectively, to allow a comparison from three  perspectives: simplicity, state-of-the-art for event activity recognition, and recognition with infrared data, the gold standard to analyze activity in  dark environments. Each baseline is built as detailed next.

\noindent \textbf{Blobs-E.}
We consider a blob-based event representation with a simple SVM classifier with polynomial kernel.  
This simple representation groups events together based on their spatial and temporal proximity, forming blobs that encapsulate regions of meaningful activity. 
Each blob is described by its center, ($c_x(t_k)$, $c_y(t_k)$),  size, ($r_{x}(t_k)$, $r_{y}(t_k)$), and weight, $\omega(t_k)$ at different times $t_k$.

Events are processed as incrementally to form blobs, assigning them to a blob using a distance criterion to the blob center. We update blob features with a convex combination of the previous blob values and the new event values. To perform the action classification we compute a constant-size feature vector of $N$ evenly time-spaced values of the blob features, from the moment it is created until it disappears due to a low weight value, i.e., each feature vector has $5\cdot N$ descriptors, describing the blob spatial and temporal evolution. Final activity recognition is done by processing the blob information with a polynomic kernel of dimension 1 and balanced class weights.

\noindent \textbf{GET.}
Peng \etal \cite{peng2023get} introduce a novel ViT backbone for event-based vision called Group Event Transformer (\textit{GET}). This model decouples temporal-polarity information from spatial information. It introduces a new event representation called Group Token, which groups asynchronous events based on their timestamps and polarities. 
Their results show how \textit{GET} outperforms all other state-of-the-art methods in various datasets of \cref{tab:datasets}.

\noindent \textbf{ResNet-IR.}
This method tries to replicate our \textit{ResNet-E} approach but for infrared data. 
Taking into account the temporal information provided by the event frames ($T_{M}$) and the frame rate (6fps) of the infrared recordings, we employ a stacking technique to cover an equivalent time window. Specifically, each original gray-scale infrared frame was stacked together with the $m=~\lfloor 6 T_{M}\rfloor -1$ previous frames. Thus, the input to the first convolutional layer are $m+1$ channels.

\subsection{Configuration details.}\label{Sec:blobbasedevent}

\indent \textbf{Pre-processing.}
All the events outside the bed area are removed beforehand. Further pre-processing steps to build each approach are detailed in Sec.~\ref{Sec:SuppConfig}.

\noindent \textbf{Training details.}
Our training uses the train set given in the dataset repository. To fine-tune the hyperparameters of our models, we extracted one subject (subject 11) from the training set as validation set. We train \textit{GET} with the  configuration shown by the authors for DVS18Gesture dataset, since this dataset is quite similar to ours (\cref{tab:datasets}). 
Regarding our approaches, we run a grid search to explore a wide range of hyperparameters, including learning rate, weight decay, and number of epochs. 
The models were optimized with standard cross entropy loss, trained with Adam optimizer.

\noindent \textbf{Evaluation details.}
The performance of each trained model is evaluated with the average top-1 accuracy of the classes (Acc@1), including the performance for each class individually, and with the classes aggregated by type, i.e., if the class corresponds to an activity with movement (\textit{Motion}) or theoretically \textit{static}.

\subsection{Classification Results}\label{Sec:Resources} 
Table~\ref{tab:AllResults}
presents the results for our approaches from \cref{Sec:Architecture} and baselines. Our 
event based approaches \textit{ResNet-E} and \textit{Dinov2-ER} obtain 
more consistent and accurate predictions, in particular for the \emph{motion} labels, while the \textit{ResNet-IR} solution distinguishes better among the \emph{static} labels. 
\textit{ResNet-E} presents the best results overall, with highest accuracy and a good compromise between accuracy and memory (\#params) and execution time requirements. 
Supplementary material (Sec.~\ref{Sec:SuppResults}) includes more detailed and visual results of the activity classification in sample recordings.

\begin{table*}[htb]
  \caption{Performance for all \textbf{Activity classification approaches}. Columns show, left to right: number of model parameters, (total and trainable); train and average inference per clip execution time; accuracy reported individually for each class, aggregated by label type (Motion-Static) and averaged for all classes. Best  results are shown in bold.}
  \label{tab:AllResults}
  \centering
  \footnotesize
  \begin{tabular}{l@{\hspace{1mm}}c@{\hspace{2mm}}c@{\hspace{3mm}}c@{\hspace{3mm}}c@{\hspace{3mm}}c@{\hspace{3mm}}c@{\hspace{3mm}}c@{\hspace{3mm}}c@{\hspace{3mm}}c@{\hspace{3mm}}c@{\hspace{3mm}}c@{\hspace{3mm}}c@{\hspace{3mm}}c@{\hspace{3mm}}c@{\hspace{3mm}}c}
    \toprule
    Approach & Params & \multicolumn{1}{c}{Time (s)} & \multicolumn{7}{c}{Acc@1 - Motion Labels} & \multicolumn{5}{c}{Acc@1 - Static Labels}  & 
    Acc@1     \\
    \cmidrule(r){1-1}  \cmidrule(r){2-2} \cmidrule(r){3-3} \cmidrule(r){4-10}
    \cmidrule(r){11-15} \cmidrule(r){16-16}
      & Total (Train) &	Train/Inf. & Head & Hands & RollL &	RollR &	Legs &	Arms & \underline{Avg.} & LieL & LieR & LieU & LieD & \underline{Avg.} &   \underline{Avg.} 
      \\
    \midrule
         \textit{GET} \cite{peng2023get} & 4.5M (4.5M) & 13980/0.02  &  0.61	& 0.66 & 1.00 & 0.97 &	1.00 &	0.91 & 0.86 &	0.62 &	0.00 &	0.88 &	0.08  & 0.40 & 0.67 \\
         \textit{Blobs-E} & 500 (500)  & 8e-2/1e-5  & 0.46 & 0.67 & 0.44 & 0.53 & 0.50 & 0.83 & 0.57 & 0.75 & 0.25 & 0.13 & 0.25 & 0.34 & 0.48  \\
         \textit{Dinov2-ER (Ours)} & 21M (3.8k)  & 16 / 0.05 &  0.84 & 0.95 & 0.58 & 0.88 & 0.91 & 0.58 & 0.80 & 0.75 & 0.58 &	0.68 & 0.58 & 0.65  & 0.73   \\
         \textit{ResNet-E (Ours)} & 11.6M (11.6M) & 263 / 0.02 & 1.00	&  0.95 & 1.00 & 0.87	& 1.00 & 1.00 & \textbf{0.97} &  0.70 & 0.58 &	0.88 & 0.25 & 0.61 & \textbf{0.82} \\
         \midrule
        \textit{ResNet-IR} & 11.6M (11.6M)  & 271 / 0.02  &  1.00  &  0.68 & 0.77 & 0.97 & 	0.66 &	0.58 & 0.78 & 0.75 & 0.66 & 0.98 &	0.33 &  \textbf{0.68} & 0.74 \\
    \bottomrule
  \end{tabular}
\end{table*}

 \noindent \textbf{Influence of scene configurations.}
The influence of different dataset room configurations (Cfg1, Cfg2 or Cfg3) is analyzed with a 3-fold validation experiment: training with data from each configuration separately and evaluating in each of the configurations separately. The results observed confirm the better performance of \textit{ResNet-E}, since it gets more robust results to changes in configuration between train and test (e.g., blanket or no-blanket).
The supplementary material (Sec.~\ref{Sec:SuppResults}) includes extended results of this experiment.

\begin{table}[t]
\caption{Accuracy, computational details and calibration metrics of the different Bayesian methods. Best results in bold, second best underlined.}
  \label{tab:Bayesian}
  \footnotesize
  \centering
  \resizebox{\columnwidth}{!}{%
  \begin{tabular}{c@{\hspace{3mm}}c@{\hspace{1mm}}c@{\hspace{1mm}}c@{\hspace{3mm}}c@{\hspace{3mm}}c@{\hspace{2mm}}c@{\hspace{2mm}}c}
    \toprule
    Approach & \multicolumn{2}{c}{Bayesian} & \multicolumn{1}{c}{Time (s)} &  \multicolumn{2}{c}{Acc@1}  & \multicolumn{2}{c}{Calibration} \\
    \cmidrule(r){1-1}  \cmidrule(r){2-3}  \cmidrule(r){4-4} \cmidrule(r){5-6} \cmidrule(r){7-8}
      & Ens. & Lap. & Train/Inf. & \textit{Mode}& \textit{Prob.} & ACE $\downarrow$ & MCE $\downarrow$ 
      \\
    \midrule
        \textit{Dinov2-ER} & - & - & 16/4e-4 &  0.73 & 0.75 &  0.10 & 0.18  \\
        \textit{Dinov2B-ER} & - & \checkmark & 25/2e-3 & 0.73 & 0.75 & 0.06 & 0.12 \\
        \textit{Dinov2B-ER} & 32 & - & 502/0.01 &  0.68 & 0.72 & 0.10 & 0.18 
        \\
        \textit{Dinov2B-ER} & 32 & \checkmark  & 927/0.06 &  0.68 & 0.72  & \underline{0.06} & \underline{0.11} 
        \\
        \textit{ResNet-E} & - & - & 263/0.02 &  \textbf{0.82} & \underline{0.81} & 0.13 & 0.20  \\
        \textit{ResNetB-E} & - & \checkmark & 271/0.03 & \textbf{0.82}  & \textbf{0.82} & 0.08 & 0.17  \\
        \textit{ResNetB-E} & 32 & - &  8429/0.64 & \underline{0.79} & 0.80 & \underline{0.06} 
        & 0.18 
        \\
        \textit{ResNetB-E} & 32 & \checkmark & 9552/0.96 & \underline{0.79} & \underline{0.81}  & \textbf{0.03} & \textbf{0.10} 
        \\
    \bottomrule
  \end{tabular}}
\end{table}

\begin{figure}[!tb]
    \centering        
       \includegraphics[width=\linewidth]{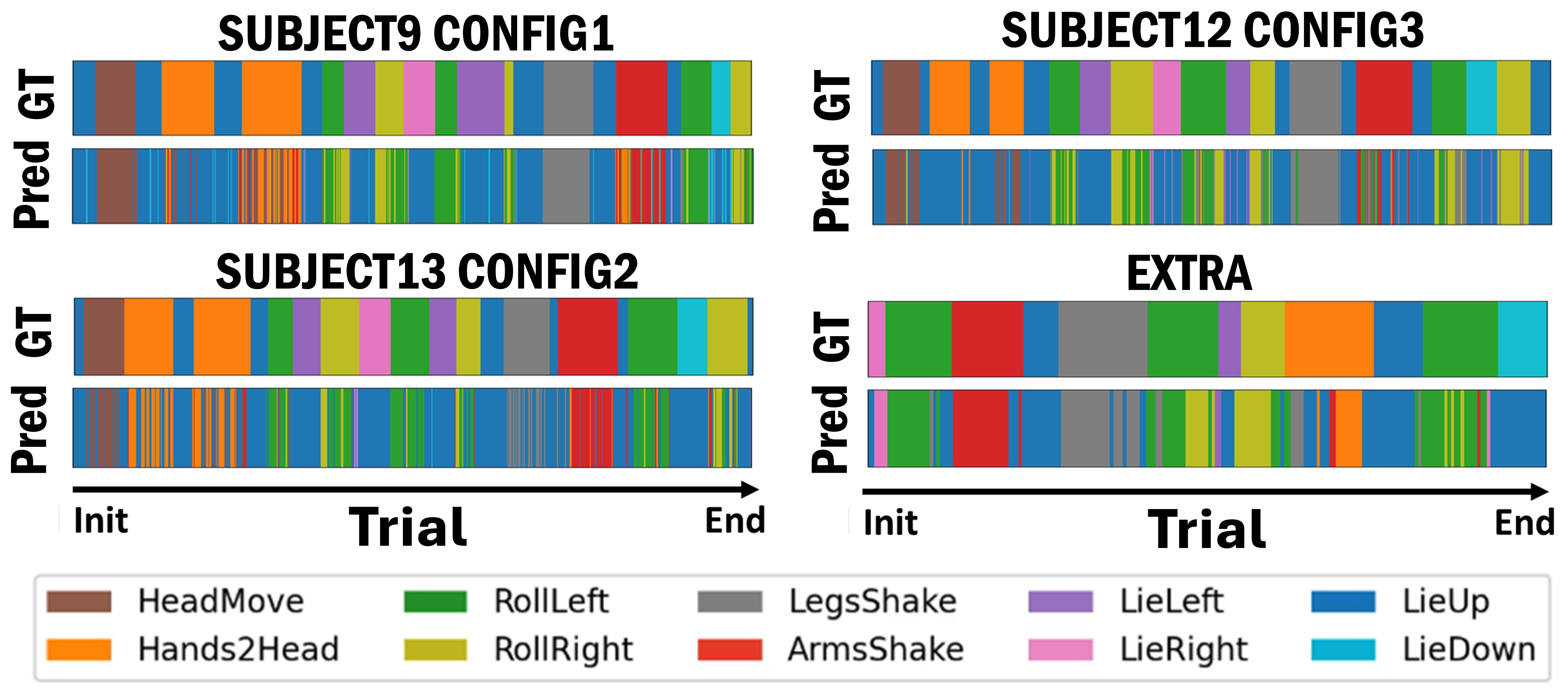}
    \captionof{figure}{\textit{ResNet-E} online qualitative results in full sequences. Each bar corresponds to a complete trial and is colored according to the clip ground truth (GT) or the frame model prediction (Pred).}
    \label{fig:online}
\end{figure}

\noindent
 \textbf{Online activity analysis.} \textit{ResNet-E} is directly applicable to the online task, by running the classification per frame instead of computing the \textit{mode} to obtain the prediction per clip. \Cref{fig:online} shows qualitatively the \textit{ResNet-E} predictions per frame (online) in full sequences (complete trial) of the test set. Quantitative results are available in 
 Sec.~\ref{Sec:SuppResults} of the supplementary material. 
 The gap in the accuracy between clip and online prediction reveals certain limitations of our current best approach. It is not capable to leverage any temporal information of the previous frame predictions to enhance the current prediction. This heuristic could be introduced with other architectures such as RNN or Transformers, and could also help to distinguish the static labels. Furthermore, not all studied methods are suitable to run the activity analysis online. By construction,  using the event reconstruction representation (that benefits from analyzing the complete sequence) prevents \textit{Dinov2-ER} from running online.

\subsection{Bayesian Methods.}\label{Sec:bayesian}
We have exhaustively evaluated the different Bayesian alternatives considered to improve the calibration and uncertainty estimation of our models.

\noindent \textbf{Calibration.}
To achieve this goal, we adopt the Reliability Diagrams \cite{degroot1983comparison}. These diagrams show the gap between confidence and accuracy per bins. Specifically, we split the predictions in $M=10$ bins according to their confidence and, for each bin $B_{m}$, we compute its average confidence $C_{m}$ and its average accuracy $A_{m}$. From these diagrams, we may compute the Average Calibration Error (ACE) and the Maximum Calibration Error (MCE) as ${\text{ACE} = \frac{1}{M^{+}}\sum_{m=1}^{M}|B_{m}-A_{m}|}$, $\text{MCE} = \max_{m}|B_{m}-A_{m}|$, 
where $M^{+}$ is the number of non-empty bins \cite{neumann2018relaxed}.

 Table~\ref{tab:Bayesian} presents a summary of these results. First, we notice that \emph{Laplace ensembles} reports the best calibration metrics because the mixture of Gaussians better captures the model uncertainty, resulting in more robust predictions. This is the first work which demonstrates on a real-world computer vision problem that \emph{Laplace ensembles} significantly outperforms both deep ensembles and Laplace approximation. For reference of deterministic performance, we have selected the networks with best accuracy. Therefore, deep ensembles, as an averaging method slightly decrease the performance with respect to the best network. However, the \textit{accumulated probability} strategy from \cref{Sec:bayesian} enables us to reach our top accuracy results. The full \emph{reliability diagrams} can be found in Sec.~\ref{Sec:SuppBayesianResults}.

\begin{figure*}[!tb]
    \centering
       \begin{subfigure}{0.32\linewidth}
        \includegraphics[width=\linewidth]{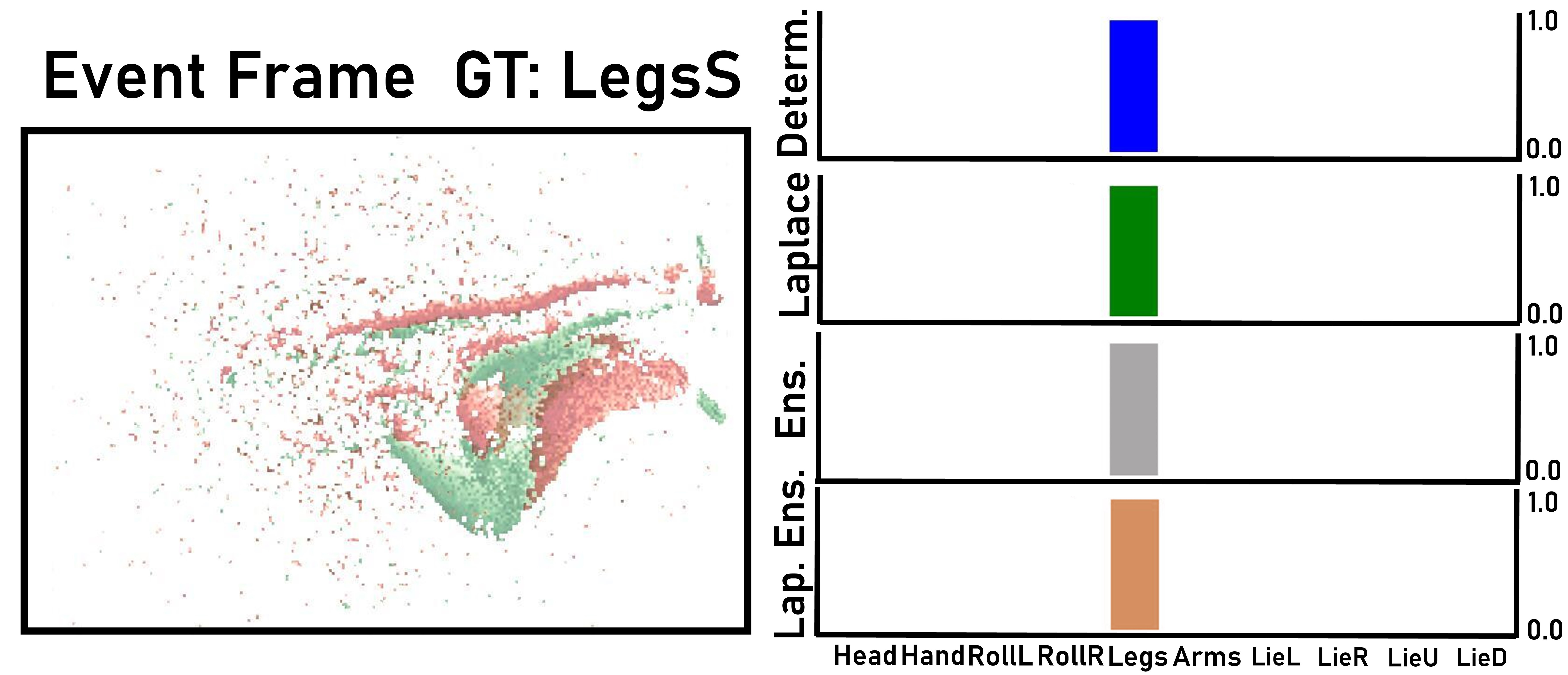}
    \end{subfigure}
    \begin{subfigure}{0.32\linewidth}
        \includegraphics[width=\linewidth]{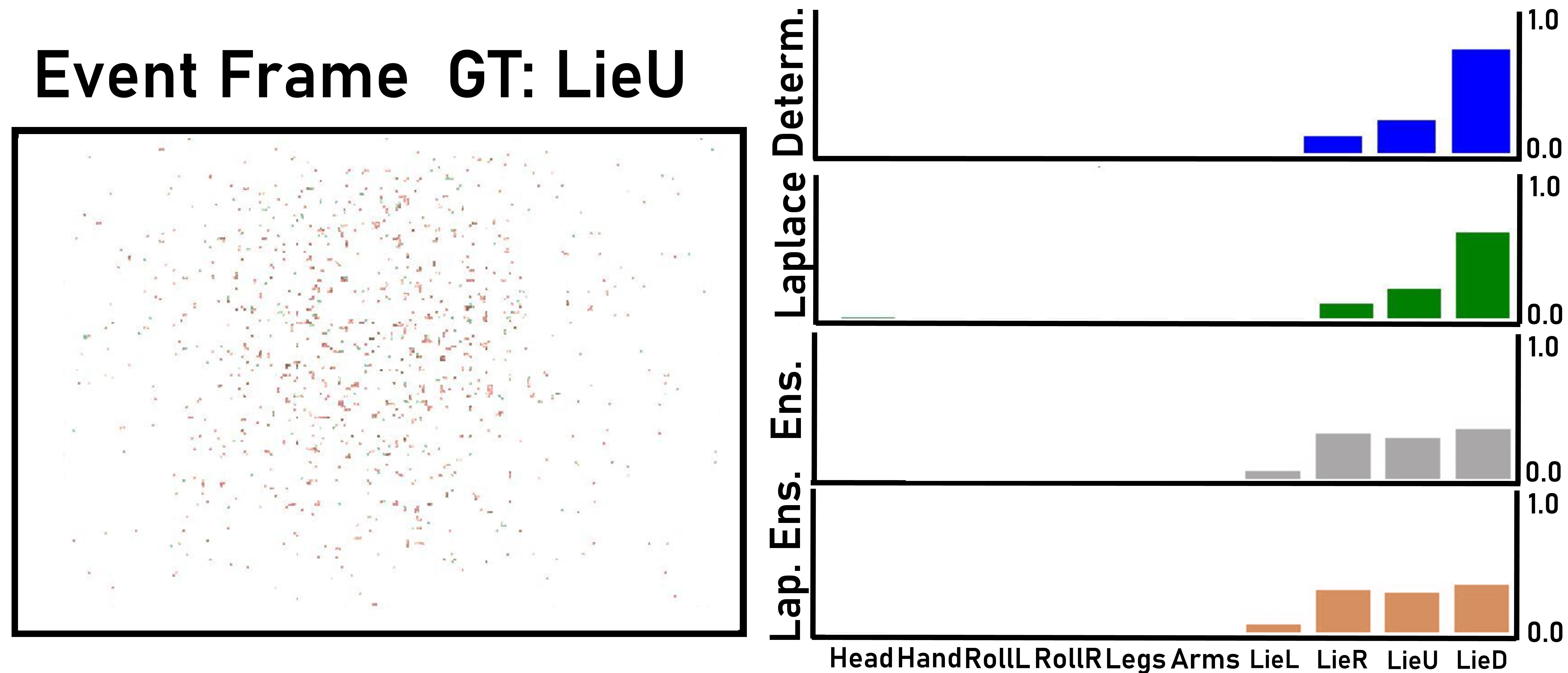}
    \end{subfigure}
    \begin{subfigure}{0.32\linewidth}
        \includegraphics[width=\linewidth]{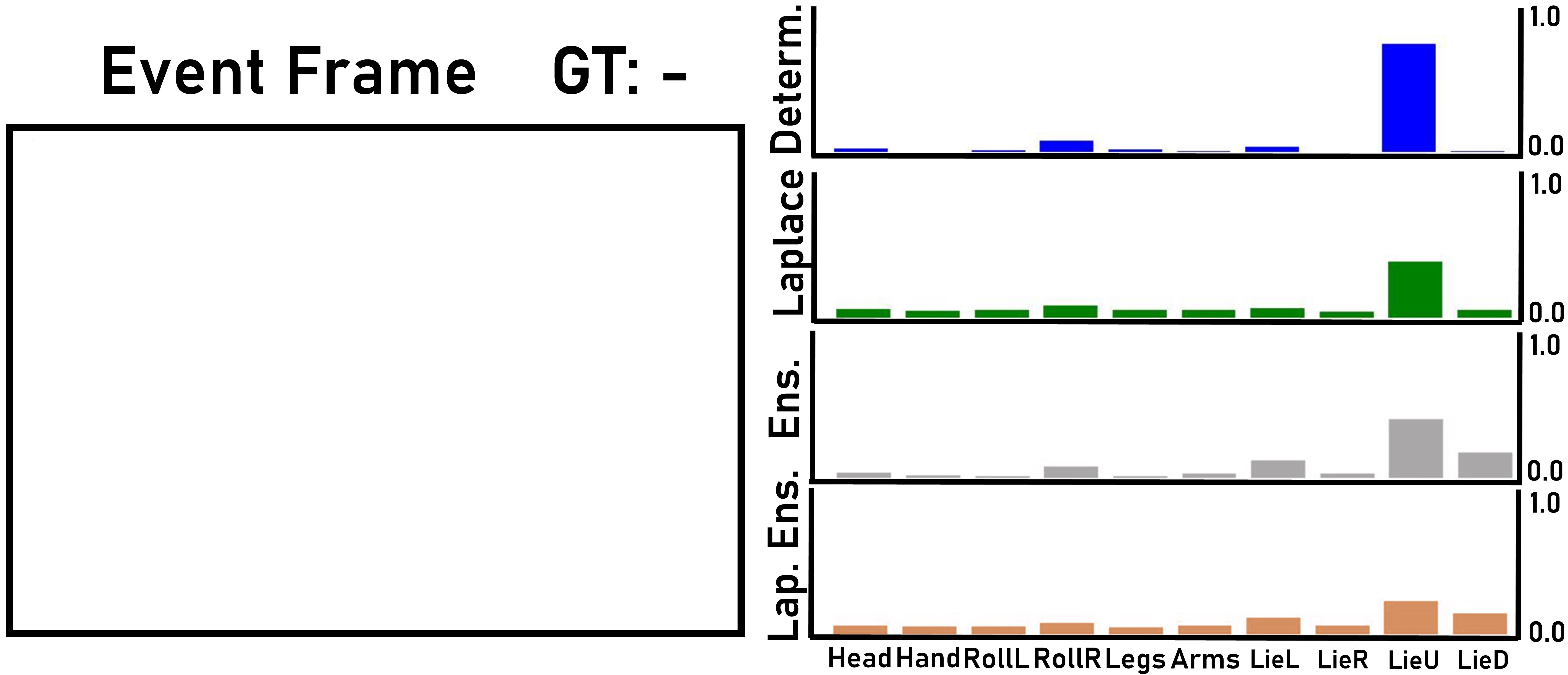}
    \end{subfigure}
    \caption{Three examples: (1) In-distribution frame where all the models predict correctly. (2) All the models misclassify the static pose frame but ensemble methods identify the uncertainty among static labels. (3) Almost empty frame where Laplace approximation reduces confidence.}
    \label{fig:OOD}
\end{figure*}

\noindent \textbf{Out-of-distribution detection.}
We additionally provide some examples (\cref{fig:OOD}) of the predictions of our different Bayesian approaches when we test them with in-distribution and out-of-distribution frames (scenario in which calibration plays a meaningful role).

\section{Conclusion}
\label{sec:conclusion}
\texttt{EventSleep} provides a valuable dataset, methods and calibration tools of event cameras in low light conditions. This type of scenarios can open many application possibilities but are lacking in existing event-data benchmarks. 
Our dataset is a unique resource that enables the development and evaluation of new techniques for sleep activity recognition and analysis, improving the performance and capabilities of automated tools in the sleep disorders diagnosis process. This use case emphasizes the benefits of event cameras under low lighting conditions.
Our proposed approach performs better in this challenging use case than relevant baselines, including the state of the art for activity recognition with event data. The experimental results obtained demonstrate the quality and usefulness of the data, and promising capabilities of the event cameras for a new application domain. 
Besides, we are the first to assess the efficacy of an unexplored Bayesian approach as \textit{Laplace ensembles} on a real-word computer vision application. 
Our results and dataset motivate the need to adapt and boost existing event-data preprocessing techniques, such as denoising or event grayscale frame reconstruction, to low-lighting scenarios. Future research directions also include the development of more sophisticated models to increase the performance or to incorporate additional information, such as the duration of the different actions in complete recordings, or more efficiency strategies, in line with the energy aware design of event camera technology.

\bibliographystyle{splncs04}
\bibliography{egbib}

\clearpage
\setcounter{page}{1}
\maketitlesupplementary
\appendix
\section{Additional dataset details}\label{Sec:SuppDataset}

\noindent
\textbf{Participants and Ethical considerations.} 
The recorded dataset includes data from 14 healthy adults converging different body shapes and sizes. The participants included 4 females and 10 males,  with heights ranging from 150cm to 190cm. The age range of the participants spans from the early twenties to late forties. All the participants were volunteers and were given written instructions and signed an informed consent form. The consent form included the authorization to make the data from the recording publicly available following the data protection regulation in Europe (General Data Protection Regulation, GDPR), which is one of the most restrictive regulations worldwide.

\section{Additional Experiments}
\subsection{Additional configuration details}\label{Sec:SuppConfig}
All the events outside the bed area are removed beforehand to eliminate noise or irrelevant background information. As result, the event grid resolution is $360 \times 500$.
Further pre-processing steps to build each approach are detailed next.

\noindent
\textbf{\textit{Blobs-E}.}
In the blob-based representation a minimum time difference of 5 ms was applied to filter the number of consecutive events in the same or nearby pixels (8-neighborhood). A second stage also filtered out events that were farther away than a time-varying threshold, $r_{x}(t_k) = \max (R_{\min},  \alpha r_{x}(t_{k-1}) +  (1 - \alpha) (c_x(t_{k-1}) - x_k)),$ with $R_{\min}=50$ a minimum distance. The hyper-parameters used are $N=100$ and $\alpha=0.9$.

\begin{figure}[!tb]
    \centering
       \begin{subfigure}{0.95\columnwidth}
        \includegraphics[width=\linewidth]{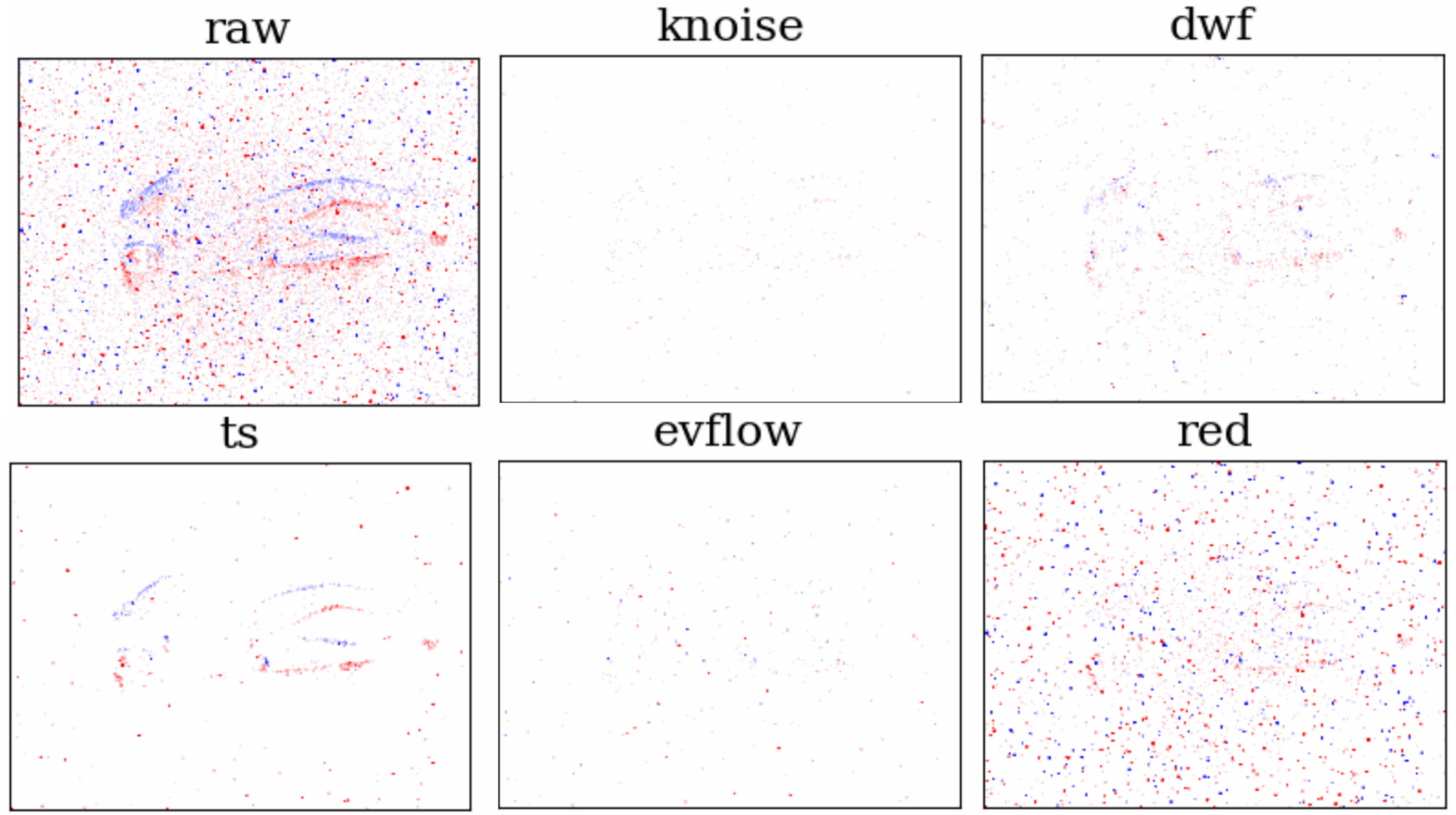}
    \end{subfigure}
    \caption{Performance comparison among different denoisers.}
    \label{fig:Denoisers}
\end{figure}

\begin{figure*}[htbp]
\centering
\rotatebox{90}{\hspace{6mm}\textbf{IR frame}} 
\includegraphics[trim=120pt 50pt 160pt 40pt, clip, width=0.19\textwidth]{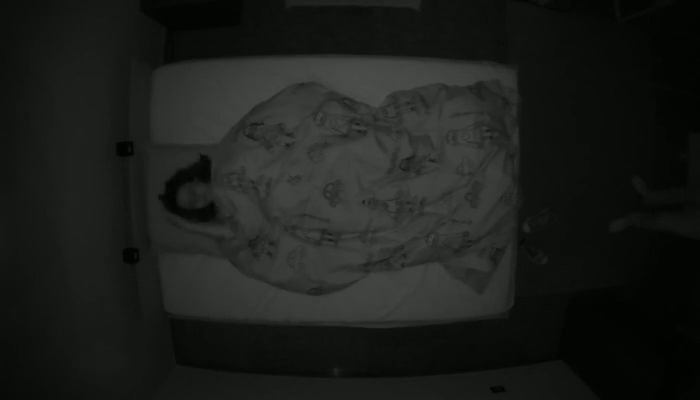}\hfill
\includegraphics[trim=120pt 50pt 160pt 40pt, clip, width=0.19\textwidth]{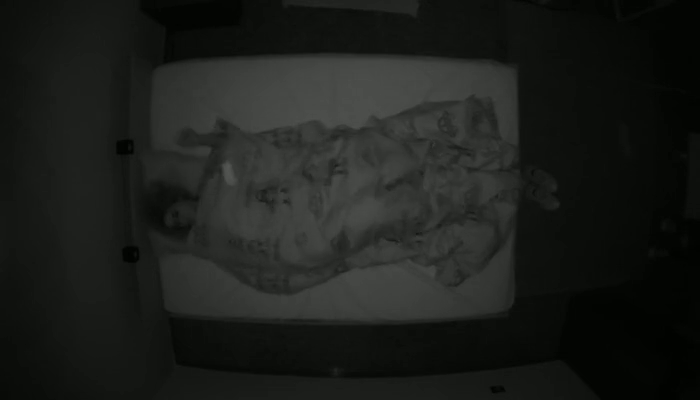}\hfill
\includegraphics[trim=120pt 50pt 160pt 40pt, clip, width=0.19\textwidth]{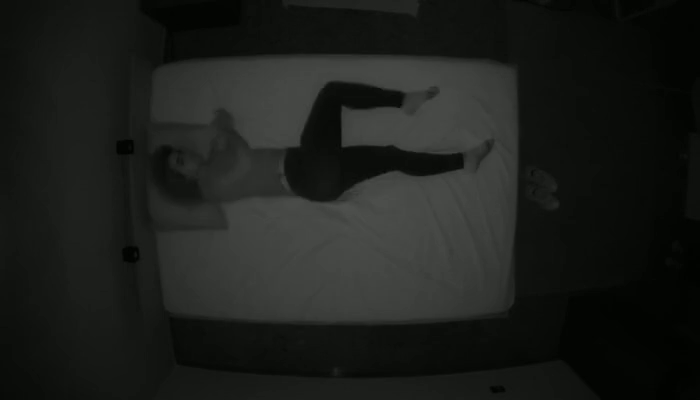}\hfill
\includegraphics[trim=120pt 50pt 160pt 40pt, clip, width=0.19\textwidth]{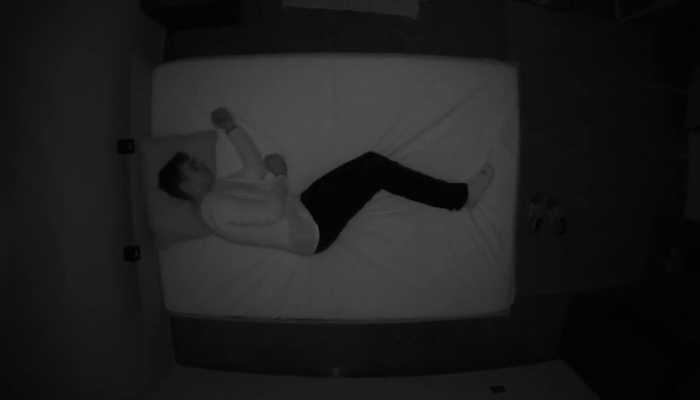}\hfill
\includegraphics[trim=120pt 50pt 160pt 40pt, clip, width=0.19\textwidth]{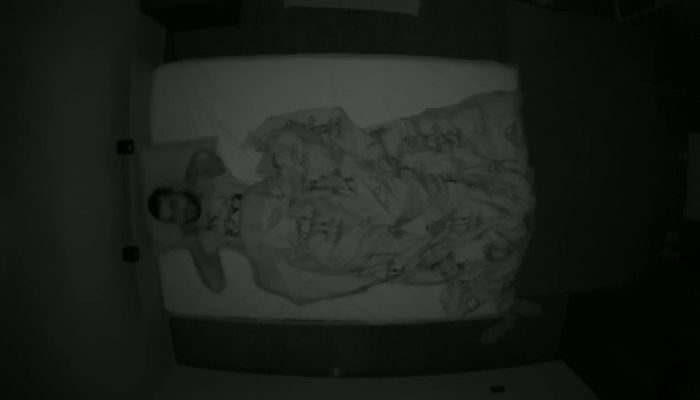}\\[0.5em]
\rotatebox{90}{\hspace{5mm}\textbf{Event Rec.}} 
\includegraphics[trim=80pt 50pt 70pt 60pt, clip, width=0.19\textwidth]{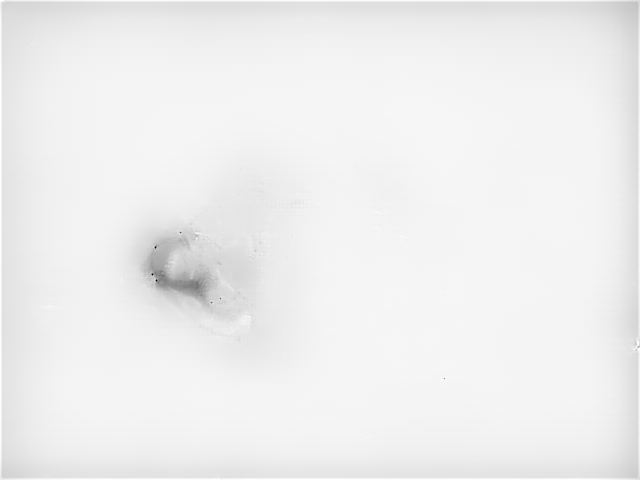}\hfill
\includegraphics[trim=80pt 50pt 70pt 60pt, clip, width=0.19\textwidth]{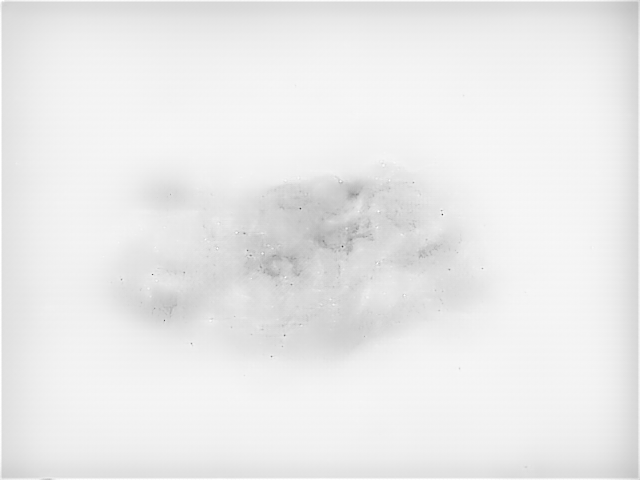}\hfill
\includegraphics[trim=80pt 50pt 70pt 60pt, clip, width=0.19\textwidth]{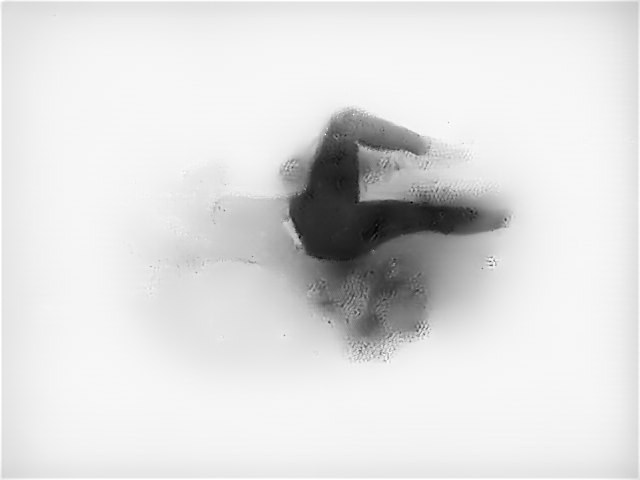}\hfill
\includegraphics[trim=80pt 50pt 70pt 60pt, clip, width=0.19\textwidth]{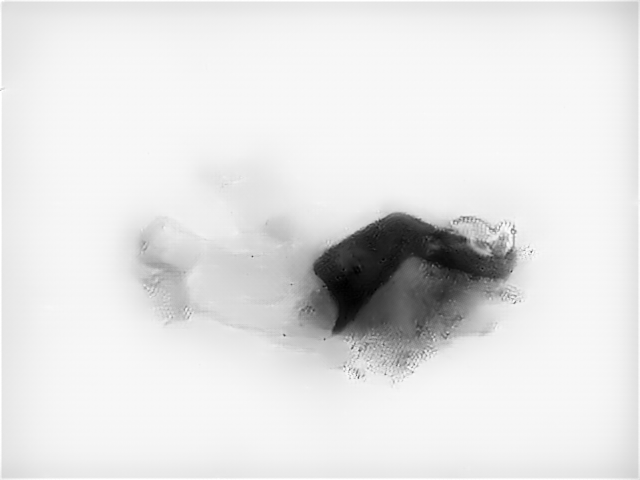}\hfill
\includegraphics[trim=80pt 50pt 70pt 60pt, clip, width=0.19\textwidth]{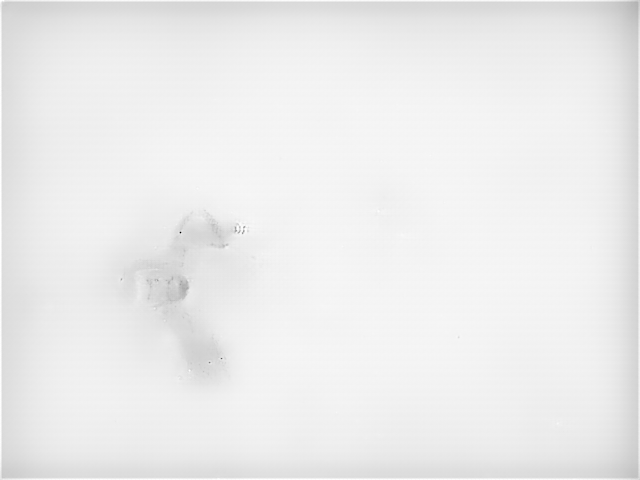}

\caption{Comparison of the level of detail that can be recovered from event recordings leveraging E2VID \cite{Rebecq19pami} with respect to frames captured by a low cost infra-red camera.}
\label{fig:mosaic_ir_events}
\end{figure*}

\noindent
\textbf{\textit{Dinov2-ER}.} For the event reconstruction representation channel, the spatio-temporal voxel grids were built with $B=5$ bins.
 Note that it was necessary to denoise the event streams beforehand to achieve proper reconstructions. Among the different methods from the E-MLB benchmark \cite{ding2023mlb}, the only one showing good denoising results in our data was the Time Surface (ts) method \cite{lagorce2016hots}. See \cref{fig:Denoisers} for an illustrative example. We provide additional examples of event reconstruction frames in \cref{fig:mosaic_ir_events}, comparing them  with respect to the temporal aligned infra-red frames.

\noindent
\textbf{\textit{ResNet-E}.} 
This approach only utilizes the 2-channel event frames $\mathcal{F}$, since we did not observe improvements adding the reconstruction channel (\cref{tab:ERvsE}). Moreover, the event reconstruction process brings large computational overload and prevents the approach from working on-line, since it requires to process the whole sequence simultaneously. The event frames representation  was achieved using $\Delta t = 0.15$s and $T_M = 0.512$s in order to achieve a frame rate similar to that of the infrared recordings and capture enough temporal information to classify the action.
We reduce the resolution by half obtaining $180\times250$ frames.

\begin{figure}[!tb]
    \centering
    \begin{subfigure}{0.48\columnwidth}
        \includegraphics[width=\linewidth]{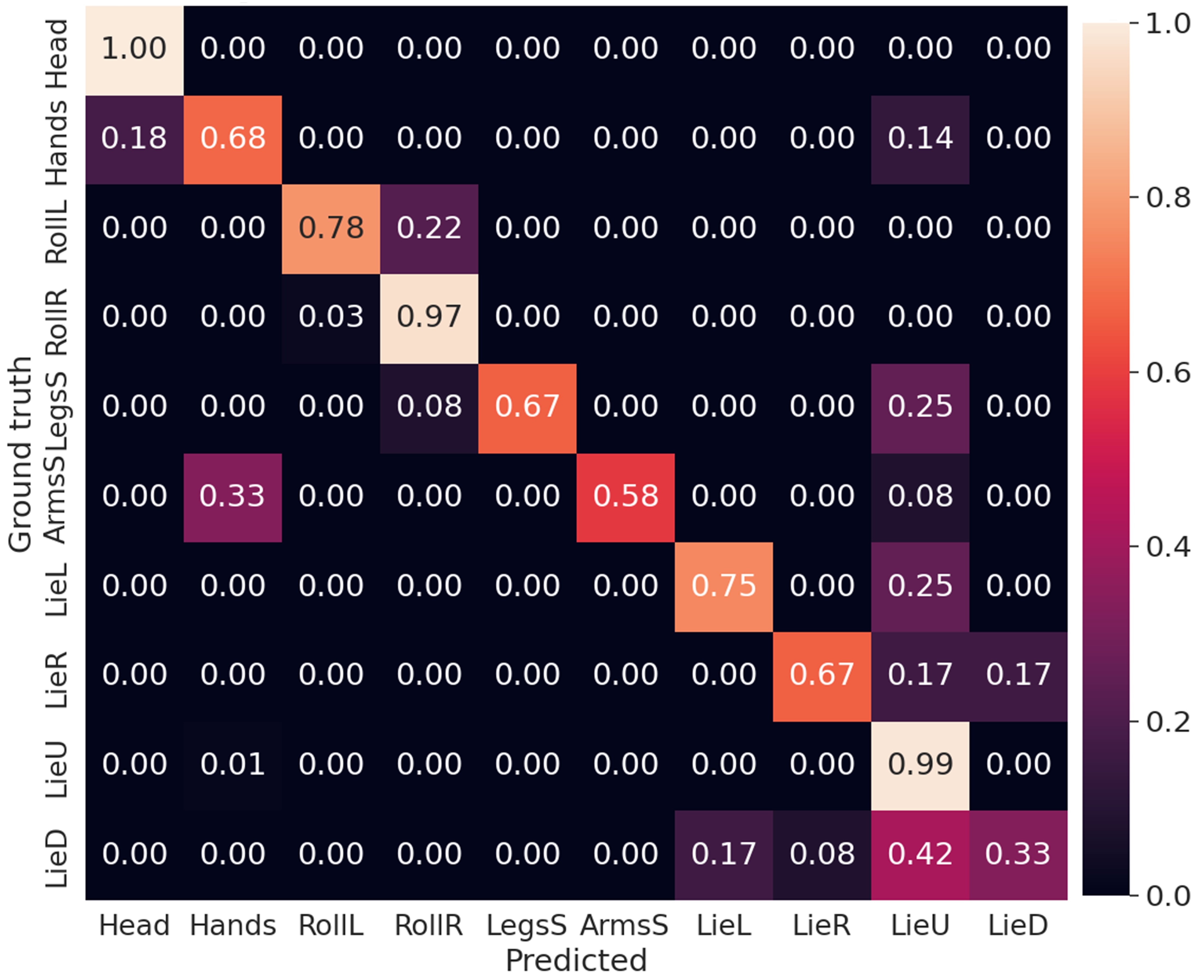}
        \caption{\textit{Dinov2-ER}.}
    \end{subfigure}
    \begin{subfigure}{0.48\columnwidth}
        \includegraphics[width=\linewidth]{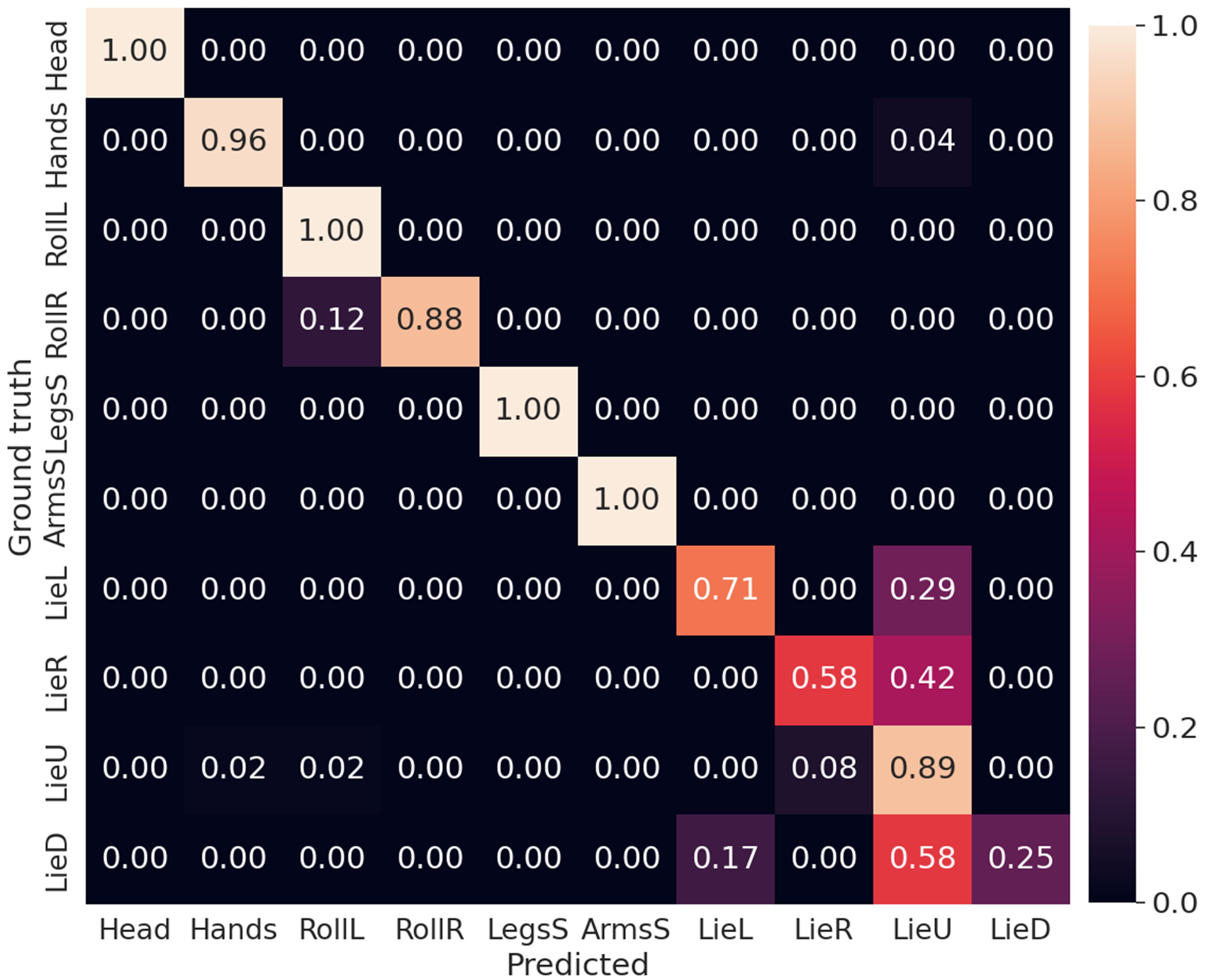}
        \caption{\textit{ResNet-E}.}
    \end{subfigure}
    \caption{Confusion matrices of our main methods.}
    \label{fig:cfm_ours}
\end{figure}

\begin{table}[htb]
  \caption{Performance of our event-based ResNet approach according to the event representation applied to its input.}
  \label{tab:ERvsE}
  \centering
  \footnotesize
  \begin{tabular}{l@{\hspace{1mm}}c@{\hspace{3mm}}c@{\hspace{3mm}}c@{\hspace{3mm}}c@{\hspace{3mm}}c@{\hspace{3mm}}c@{\hspace{3mm}}c}
    \toprule
    Approach& \multicolumn{2}{c}{Events Repr.} &  Online  & \multicolumn{3}{c}{Acc@1} \\
    
    \cmidrule(r){1-1} \cmidrule(r){2-3}  \cmidrule(r){4-4} \cmidrule(r){5-7}
    & Frame & Reconst. & & Motion & Static & AVG.
      \\
      \midrule
         \textit{ResNet-ER} & \checkmark & \checkmark & \xmark  
         & 0.95	&  0.55 & 0.79 \\
        \textit{ResNet-E} & \checkmark & \xmark & \checkmark 
        & \textbf{0.97}  
        & \textbf{0.61} & \textbf{0.82} \\
    \bottomrule
  \end{tabular}
\end{table}

\noindent
\textbf{\textit{ResNet-IR}.} Since $T_{M}=0.512$s, we stack each frame with the previous $m=2$ frames, achieving 3-channel frames. We also reduced the resolution to $180\times250$.

\subsection{Additional classification results}\label{Sec:SuppResults}

\begin{table*}[!tb]
  \caption{Accuracy of our proposed approaches (the best performing in the overall tests) reported individually for each configuration. 
  }
  \label{tab:Configs}
  \centering
  \footnotesize
  \resizebox{\textwidth}{!}{%
  \begin{tabular}{l c c c c c c c c c c c c c c c c c}
    \toprule
    TrainConfig & \multicolumn{4}{c}{Config1} & \multicolumn{4}{c}{Config2}   & \multicolumn{4}{c}{Config3}  & \multicolumn{4}{c}{AllConfigs}       \\
    \cmidrule(r){1-1}  \cmidrule(r){2-5}
    \cmidrule(r){6-9} \cmidrule(r){10-13} \cmidrule(r){14-17} 
       TestConfig & Cfg1 & Cfg2 & Cfg3 & All & Cfg1 & Cfg2 & Cfg3 & All & Cfg1 & Cfg2 & Cfg3 & All & Cfg1 & Cfg2 & Cfg3 & All \\
       
    \midrule
        \textit{Dinov2-ER} & 0.65 & 0.64 & 0.56 & 0.62 & 0.67 & 0.64 & 0.45 & 0.59 & 0.60 & 0.60 & 0.70 & 0.63 & 0.70 & 0.75 & 0.76 & 0.73 \\
         \textit{ResNet-E} & 0.70 & 0.68 & 0.72 & 0.70 & 0.67 & 0.79 & 0.52  & 0.67 & 0.60 & 0.70 & 0.72 & 0.68 &  0.79 & 0.84 & 0.85 & 0.82 \\
    \bottomrule
  \end{tabular}}
\end{table*}

\noindent \textbf{Hardware used.}
The experiments were conducted using a  Intel Core™ i7-12700K processor with 20 cores, and NVIDIA GeForce RTX 3090 GPU.

\noindent \textbf{Confusion matrices.}
We provide confusion matrices to further analyze the performance of our models (\cref{fig:cfm_ours}). These figures highlight the challenges faced by our models in accurately distinguishing between the ``static'' labels. Some confusion can also be found between the two roll classes. Our model based on Dino also suffer some misclassification between the \textit{Hands2Head} and \textit{ArmsShake} classes, which can be attributed to their inherent similarity.

\noindent \textbf{Influence of scene configurations.} 
The influence of different dataset room configurations (Cfg1, Cfg2 or Cfg3) is analyzed with a 3-fold validation experiment: training with data from each configuration separately and evaluating in each of the configurations separately. The results in \cref{tab:Configs} confirm the better performance of \textit{ResNet-E}, since it gets more robust results to changes in configuration between train and test (e.g., blanket or no-blanket).

\noindent \textbf{Online classification.} We define online classification as the task of classifying an event frame as soon as it is available, instead of performing per clip statistics to provide a prediction. Our model of \textit{ResNet-E} can be directly applied to the online task, as it can provide a prediction directly by frame. \cref{fig:online} in the main document shows qualitatively the \textit{ResNet-E} predictions per frame (online) in full sequences (complete trial) of the test set. Since online predictions are based on less information, the accuracy of our \textit{ResNet-E} model drops with respect to clip-based prediction (\cref{tab:Online}).

\begin{table}[tb]
  \caption{Performance of \emph{ResNet-E} at different temporal windows. Best  results in bold.}
  \label{tab:Online}
  \centering
  \footnotesize
  \begin{tabular}{l@{\hspace{1mm}}c@{\hspace{3mm}}c@{\hspace{3mm}}c@{\hspace{3mm}}c@{\hspace{3mm}}c}
    \toprule
    Approach & Temp. Window  & \multicolumn{3}{c}{Acc@1} \\
    \cmidrule(r){1-1}  \cmidrule(r){2-2} \cmidrule(r){3-5}
    & & Motion & Static & AVG.
      \\
      \midrule
         \textit{ResNet-E (Ours)} & Clip  
         & \textbf{0.97} 
         & \textbf{0.61} & \textbf{0.82} \\
        \textit{ResNet-E (Ours)} & Frame (0.5s)   
        & 0.77 
        & 0.48 & 0.66 \\
    \bottomrule
  \end{tabular}
\end{table}

\subsection{Additional Bayesian methods results}\label{Sec:SuppBayesianResults}
\begin{figure}[!t]
    \centering
       \begin{subfigure}{0.49\columnwidth}
        \includegraphics[width=\linewidth]{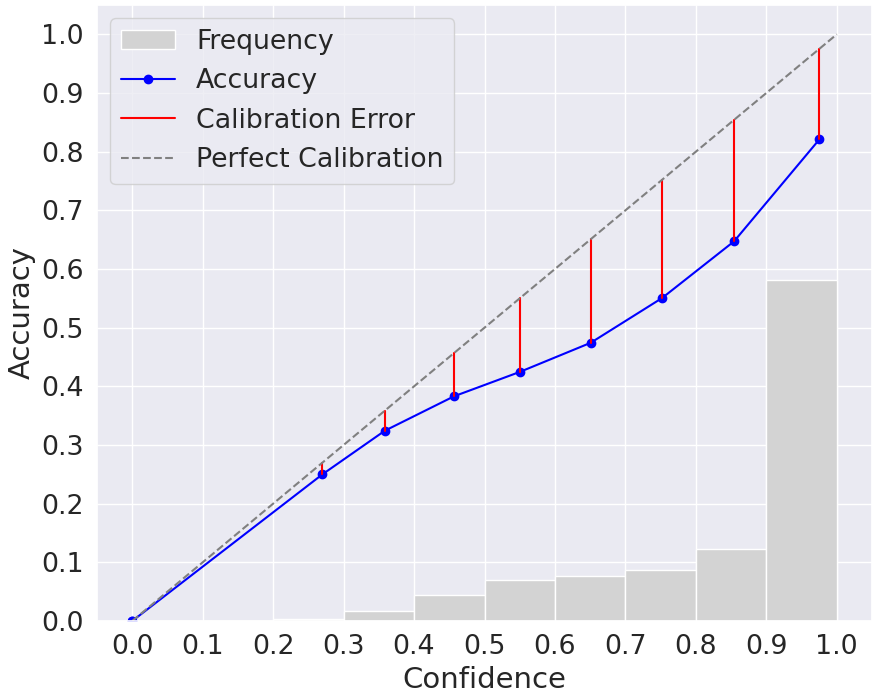}
        \caption{\textit{ResNet-E}.}
    \end{subfigure}
    \begin{subfigure}{0.49\columnwidth}
        \includegraphics[width=\linewidth]{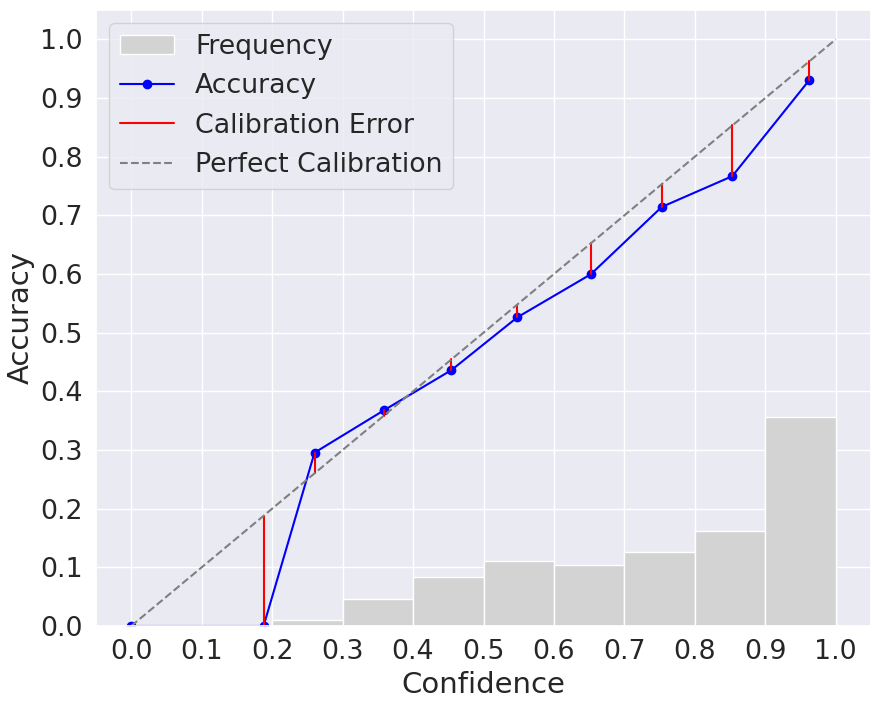}
        \caption{\textit{ResNetB-E} with ensembles.}
    \end{subfigure}
    \begin{subfigure}{0.49\columnwidth}
        \includegraphics[width=\linewidth]{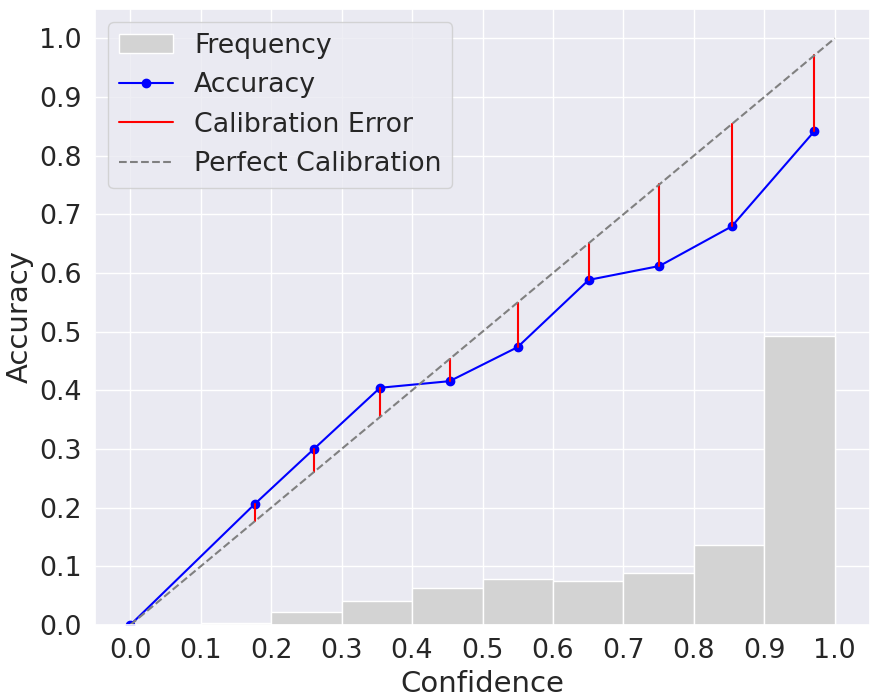}
        \caption{\textit{ResNetB-E} with Laplace.}
    \end{subfigure}
        \begin{subfigure}{0.49\columnwidth}
        \includegraphics[width=\linewidth]{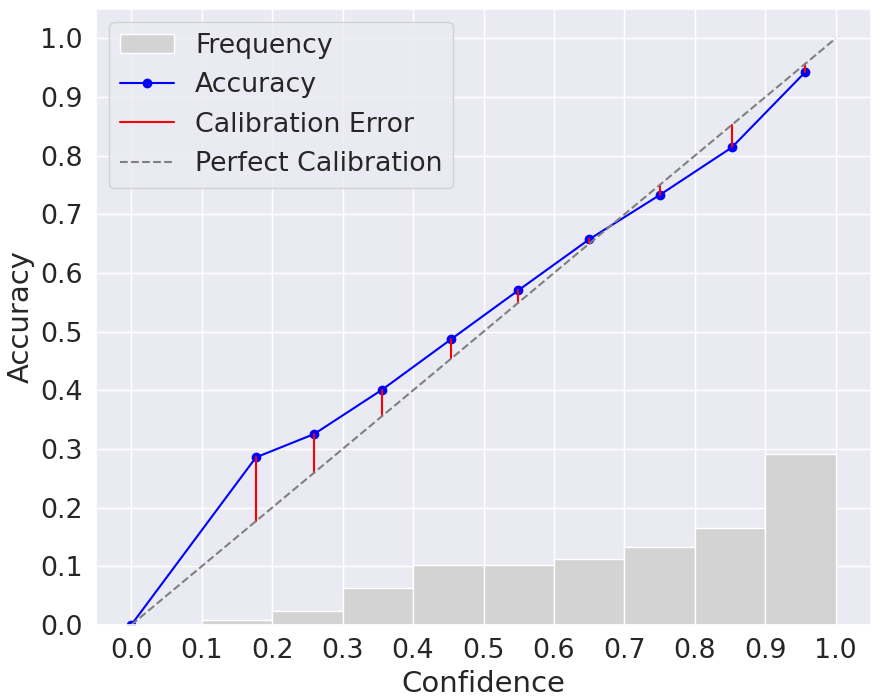}
        \caption{\textit{ResNetB-E} with \textit{Lap. ens.}}
    \end{subfigure}
    \caption{Reliability diagrams plot the accuracy vs the predicted confidence respect to a perfectly calibrated model. \textit{\textit{Laplace ensembles}} achieves the best calibration. }
    \label{fig:Reliability}
\end{figure}

\textbf{Calibration.}
With the aim of qualitatively measuring the difference between accuracy and confidence, we use the Reliability Diagrams \cite{degroot1983comparison}. Regarding \cref{fig:Reliability}, \emph{Laplace ensembles} approach achieves the best calibration for \emph{ResNet-E} method.

\end{document}